\documentclass{article}

    \PassOptionsToPackage{numbers, compress}{natbib}
    \usepackage[final]{neurips_2020}

\usepackage[utf8]{inputenc} % allow utf-8 input
\usepackage[T1]{fontenc}    % use 8-bit T1 fonts
\usepackage{hyperref}       % hyperlinks
\usepackage{url}            % simple URL typesetting
\usepackage{booktabs}       % professional-quality tables
\usepackage{amsfonts}       % blackboard math symbols
\usepackage{nicefrac}       % compact symbols for 1/2, etc.
\usepackage{microtype}      % microtypography

\usepackage{algorithm}
\usepackage{algorithmic}
\usepackage{amsmath}
\usepackage{mathabx}
\usepackage{xspace} 
\usepackage{graphicx}
\usepackage{subfig}
\usepackage{xcolor}
\usepackage{xparse}
\usepackage{float}
\usepackage{enumitem}
\usepackage{euscript}
\usepackage{soul}
\usepackage{tabulary}
\usepackage{multirow}
\usepackage{capt-of}
\usepackage{makecell}
\usepackage[capitalize]{cleveref}
\usepackage{mathrsfs}
\usepackage{wrapfig}

\NewDocumentCommand{\hugo}{ mO{} }{\textcolor{blue}{\textsuperscript{\textit{Hugo}}\textsf{\textbf{\small[#1]}}}}

\newcommand{\alg}{\textsc{IDGL}\xspace}
\newcommand{\salg}{\textsc{IDGL-Anch}\xspace}

\let\vec\mathbf

\title{Iterative Deep Graph Learning for Graph Neural Networks: Better and Robust Node Embeddings}

\author{%
  Yu Chen\\
  Rensselaer Polytechnic Institute\\
  \texttt{cheny39@rpi.edu} \\
   \And
   Lingfei Wu\thanks{Corresponding author.}\\
   IBM Research \\
   \texttt{lwu@email.wm.edu} \\
   \AND
   Mohammed J. Zaki \\
   Rensselaer Polytechnic Institute \\
   \texttt{zaki@cs.rpi.edu} \\
}

\begin{document}

\maketitle

\begin{abstract}
In this paper, we propose an end-to-end graph learning framework, namely \textbf{I}terative \textbf{D}eep \textbf{G}raph \textbf{L}earning (\alg), for jointly and iteratively learning graph structure and graph embedding. The key rationale of \alg is to learn a better graph structure based on better node embeddings, and vice versa (i.e., better node embeddings based on a better graph structure). Our iterative method dynamically stops when the learned graph structure approaches close enough to the graph optimized for the downstream prediction task. In addition, we cast the graph learning problem as a similarity metric learning problem and leverage adaptive graph regularization for controlling the quality of the learned graph. Finally, combining the anchor-based approximation technique, we further propose a scalable version of \alg, namely \salg, which significantly reduces the time and space complexity of \alg without compromising the performance. Our extensive experiments on nine benchmarks show that our proposed \alg models can consistently outperform or match the state-of-the-art baselines. Furthermore, \alg can be more robust to adversarial graphs and cope with both transductive and inductive learning. 
\end{abstract}

\section{Introduction}

% graph neural networks and applications
Recent years have seen a significantly growing amount of interest in graph neural networks (GNNs), especially on efforts devoted to developing more effective GNNs for node classification 
\citep{kipf2016semi,li2016gated,GraphSage:hamilton2017inductive,velivckovic2017graph}, 
graph classification 
\citep{ying2018hierarchical,ma2019graph} and graph generation
\citep{samanta2018designing,li2018learning,you2018graphrnn}. 
Despite GNNs' powerful ability in learning expressive node embeddings, unfortunately, they can only be used when graph-structured data is available. Many real-world applications naturally admit network-structured data (e.g., social networks). 
However, these intrinsic graph-structures are not always optimal for the downstream tasks. This is partially because the raw graphs were constructed from the original feature space, which may not reflect the ``true" graph topology after feature extraction and transformation.
Another potential reason is that real-world graphs are often noisy 
or even incomplete 
due to the inevitably error-prone data measurement or collection. 
%Some previous works~\citep{velivckovic2017graph} mitigate this issue by reweighting the importance of neighboring nodes using self-attention on node embeddings, which still assumes that the original graph connectivity information is noiseless.  
Furthermore, many applications such as those in natural language processing 
\citep{chen2019reinforcement,xu2018graph2seq,xu2018exploiting} 
may only have sequential data or even just the original feature matrix, requiring additional graph construction from the original data matrix.

To address these limitations, we propose an end-to-end graph learning framework, namely \textbf{I}terative \textbf{D}eep \textbf{G}raph \textbf{L}earning (\alg), for jointly
and iteratively learning the graph structure and the GNN parameters that are optimized toward the downstream prediction task. 
The key rationale of our \alg framework is to learn a better graph structure based on better node embeddings, and at the same time, to learn better node embeddings based on a better graph structure. 
In particular, \alg is a novel iterative method that aims to search for an implicit graph structure that augments the initial graph structure (if not available we use a kNN graph) with the goal of optimizing the graph for downstream prediction tasks. 
The iterative method adjusts when to stop in each mini-batch when the learned graph structure approaches close enough to 
the graph optimized for the downstream task.
% the optimized graph based on our proposed stopping criterion.

Furthermore, we present a graph learning neural network that uses multi-head self-attention with epsilon-neighborhood sparsification for constructing a graph. 
Moreover, unlike the work in \citep{kalofolias2016learn} that directly optimizes an adjacency matrix without considering the downstream task, we learn a graph metric learning function by optimizing a joint loss combining both task-specific prediction loss and graph regularization loss.
Finally, we further propose a scalable version of our \alg framework, namely \salg, by combining the anchor-based approximation technique,
which reduces the time and memory complexity from quadratic to linear with respect to the numbers of graph nodes.
% which learns a node-anchor affinity matrix instead of explicitly computing an all-to-all similarity matrix for all the nodes. 
% Because the number of anchors is usually much smaller than the number of nodes, we improve the time (or space) complexity from $n^2$ (or $n^2$) to $ns^2$ (or $ns$) where $n$ and $s$ are the numbers of nodes and anchors, respectively.

In short, we summarize the main contributions as follows:
% \vspace{-0.1in}
\setlist{nolistsep}
\begin{itemize}[noitemsep,topsep=0pt]
    \item We propose a novel end-to-end graph learning framework (\alg) for jointly and iteratively learning the graph structure and graph embedding. \alg dynamically stops when the learned graph structure approaches the optimized graph (for prediction).
    To the best of our knowledge, we are the first to introduce iterative learning for graph structure learning.

    % \item We cast the graph structure learning problem as a graph similarity metric learning problem and leverage adaptive graph regularization for controlling the quality of the learned graph.
    
    \item Combining the anchor-based approximation technique, we further propose a scalable version of \alg, namely \salg, which achieves linear complexity in both computational time and memory consumption with respect to the number of graph nodes.
    
    \item Experimental results show that our models consistently outperform or match the state-of-the-art baselines on various downstream tasks. 
    More importantly, \alg can be more robust to adversarial graph examples and can cope with both transductive and inductive learning. 
    
\end{itemize}

% The main contributions of our approach are: i) 
% We propose a novel end-to-end graph learning framework (\alg) for jointly learning the graph structure and graph embedding. \alg dynamically stops when the learned graph structure approaches the optimal graph (for prediction);
% %MJZ -- check what I wrote. I do not like the phrase "optimal graph". Better not to state that and open the work to bias or obvious criticism. We CANNOT know that the graph is optimal is the usual sense. It may not even be optimal for the prediction task. So I do not see how we can claim any optimality. 
% ii) We cast the graph structure learning problem as a graph similarity metric learning problem and leverage adaptive graph regularization for controlling smoothness, connectivity and sparsity of the generated graph;
% and iii) Our experiments demonstrate that \alg consistently outperforms or matches state-of-the-art baselines on various downstream tasks. 
% %More importantly, \alg can be more robust to adversarial graph examples and can cope with both transductive  and inductive learning. 
    
% %MJZ -- the point about transductive and inductive has already been mentioned previously, so I commented it out. We should not needlessly repeat the same claim in the same section. It is OK to repeat later in discussion or conclusion.

\section{Iterative Deep Graph Learning Framework}

\begin{figure}[tb]
%  \vspace{-4mm}
  \centering
    \includegraphics[keepaspectratio=true,scale=0.18]{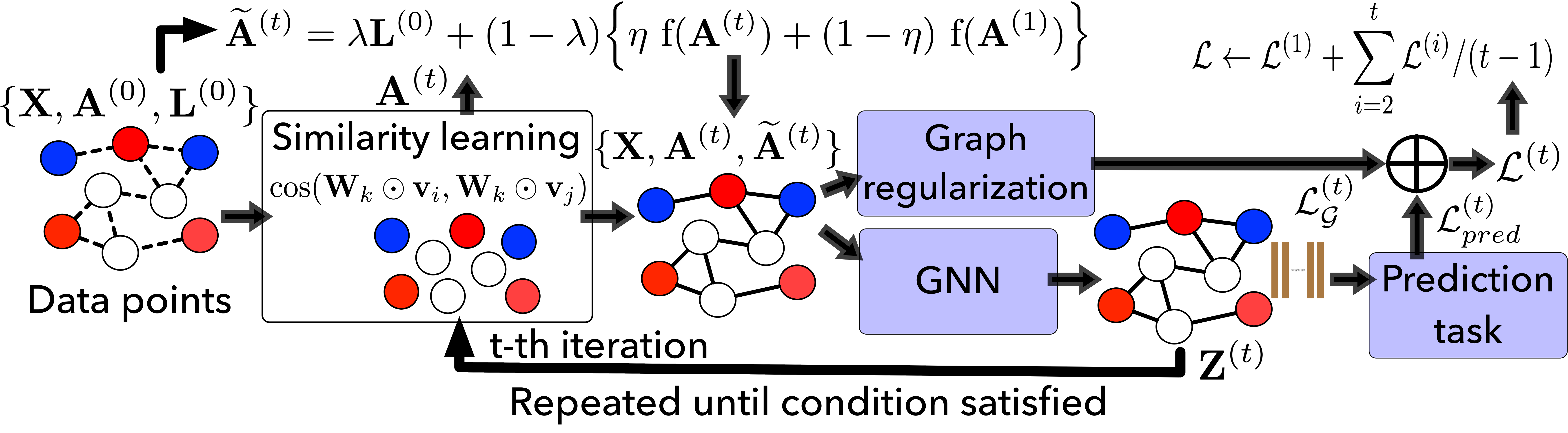}
  \caption{Overall architecture of the proposed \alg framework. Dashed lines (in data points on left) indicate the initial noisy graph topology $\vec{A}^{(0)}$ 
  (if not available we use a kNN graph). 
%   either from the original data graph or from the graph constructed using the kNN strategy. 
  }
  \label{fig:overall_arch}
 \vspace{-4mm}
\end{figure}

\subsection{Problem Formulation}
Let the graph $\mathcal{G} = (\mathcal{V}, \mathcal{E})$ be represented as a set of $n$ nodes $v_i \in \mathcal{V}$ with an initial node feature matrix $\vec{X} \in \mathbb{R}^{d \times n}$, edges $(v_i, v_j) \in \mathcal{E}$ (binary or weighted) formulating an initial noisy adjacency matrix $\vec{A}^{(0)} \in \mathbb{R}^{n \times n}$, and a degree matrix $\vec{D}^{(0)}_{ii} = \sum_j \vec{A}^{(0)}_{ij}$. Given a noisy graph input $\mathcal{G} := \{\vec{A}^{(0)}, \vec{X}\}$ or only a feature matrix $\vec{X} \in \mathbb{R}^{d \times n}$, the deep graph learning problem we consider in this paper is to produce an optimized graph $\mathcal{G}^{*} := \{\vec{A}^{(*)}, \vec{X}\}$ and its corresponding graph node embeddings $\vec{Z} = f(\mathcal{G}^{*}, \theta) \in \mathbb{R}^{h \times n}$, with respect to some (semi-)supervised downstream task. 
It is worth noting that we assume that the graph noise is only from graph topology (the adjacency matrix) and the node feature matrix $\vec{X}$ is noiseless. The more challenging scenario where both graph topology and node feature matrix are noisy, is part of our future work.
Without losing the generality, in this paper, we consider both node-level and graph-level prediction tasks.
% Specifically, we train a deep graph learning model based on a set of training pairs of graph-structured input and scalar output $\big\{(\mathcal{G}_1, y_1),...,(\mathcal{G}_n, y_n)\big \}$ drawn from some fixed but unknown probability distribution.

\subsection{Graph Learning and Graph Embedding: A Unified Perspective}
% \hugo{many overlaps with the introduction section.}

Graph topology is crucial for a GNN to learn expressive graph node embeddings. Most of existing GNN methods simply assume that the input graph topology is perfect, which is not necessarily true in practice since real-world graphs are often noisy or incomplete. More importantly, the provided input graph(s) may not be ideal for the supervised downstream tasks since most of raw graphs are constructed from the original feature space which may fail to reflect the ``true'' graph topology after high-level feature transformations. 
Some previous works~\citep{velivckovic2017graph} mitigate this issue by 
% learning new node embeddings by
reweighting the importance of neighborhood node embeddings using self-attention on previously learned node embeddings, which still assumes that the original graph connectivity information is noiseless.  

% \begin{figure}
\begin{wrapfigure}{r}{2in}
    % \vspace{-0.1in}
  \centering
  \includegraphics[width=2.0in]{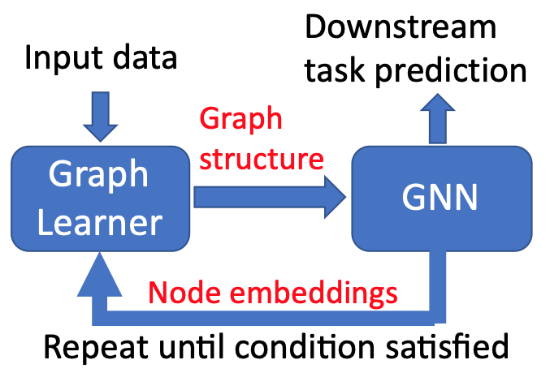}
    % \vspace{-0.2in}
  \caption{A sketch of the proposed \alg framework.}
  \label{fig:idgl_sketch}
    % \vspace{-0.1in}
\end{wrapfigure}
% \end{figure}

To handle potentially noisy input graph, we propose our novel \alg framework that formulates the problem as an iterative learning problem which jointly learns the graph structure and the GNN parameters. The key rationale of our \alg framework is to learn a better graph structure based on better node embeddings, and in the meanwhile, to learn better node embeddings based on a better graph structure,
as shown in \cref{fig:idgl_sketch}.
Unlike most existing methods that construct graphs based on raw node features, the node embeddings learned by GNNs (optimized toward the downstream task) could provide useful information for learning better graph structures. On the other hand, the newly learned graph structures could be a better graph input for GNNs to learn better node embeddings.

In particular, \alg is a novel iterative method that aims to search for an implicit graph structure that augments the initial graph structure (if not available we use a kNN graph) for downstream prediction tasks. The iterative method dynamically stops in each mini-batch when the learned graph structure approaches close enough to the optimized graph (with respect to the downstream task) based on our proposed stopping criterion. Moreover, the process of graph construction can be optimized toward the downstream task in an end-to-end manner.

\subsection{Graph Learning as Similarity Metric Learning}

Previous methods (e.g., \citep{franceschi2019learning}) that model the graph learning problem as learning a joint discrete probability distribution on the edges of the graph have shown promising performance. However, since they optimize the edge connectivities by assuming that the graph nodes are known, they are unable to cope with the inductive setting (with new nodes during testing). To overcome this issue, we cast the graph structure learning problem as similarity metric learning, which will be jointly trained with the prediction model dedicated to a downstream task.

% In graph theory, an adjacency matrix is used to represent a finite graph, where the elements of the matrix indicate the existence and strength of connections between pairs of nodes in the graph.
% In machine learning, metric learning is the task of learning a distance function over objects where the distance function is usually defined based on the similarity between pairs of objects.
% In this paper, 
% we assume that the ``semantic'' relations among objects contain certain information about the underlying graph structure,
% and formulate the graph learning problem as a similarity metric learning problem.

% \subsubsection{Graph Similarity Metric Learning}
\textbf{{Graph similarity metric learning.}}
Common options for metric learning include cosine similarity~\citep{nguyen2010cosine,wojke2018deep}, radial basis function (RBF) kernel~\citep{yeung2007kernel,li2018adaptive} and attention mechanisms~\citep{vaswani2017attention,jiang2019semi}.
A good similarity metric function is supposed to be learnable and expressively powerful.
Although our framework is agnostic to various similarity metric functions, without loss of generality, we design a weighted cosine similarity as our metric function, $s_{ij} = \text{cos}(\vec{w} \odot \vec{v}_i, \vec{w} \odot \vec{v}_j)$,
% \begin{equation}\label{eq:weighted_cosine}
% \begin{aligned}
% s_{ij} = \text{cos}(\vec{w} \odot \vec{v}_i, \vec{w} \odot \vec{v}_j)
% \end{aligned}
% \end{equation}
where $\odot$ denotes the Hadamard product,
and $\vec{w}$ is a learnable weight vector which has the same dimension as the input vectors $\vec{v}_i$ and $\vec{v}_j$, and learns to highlight different dimensions of the vectors. Note that the two input vectors could be either raw node features or computed node embeddings.

% \noindent\textbf{Multi-head similarity metric learning.}
To stabilize the learning process and increase the expressive power, we 
extend our similarity metric function
% ~\cref{eq:weighted_cosine} 
to a multi-head version (similar to the observations in \citep{vaswani2017attention,velivckovic2017graph}).
Specifically, we use $m$ weight vectors (each one representing one perspective) to compute $m$ independent similarity matrices using the above similarity function
%~\cref{eq:weighted_cosine},
and take their average as the final similarity:
\begin{equation}\label{eq:similarity_metric_learning}
\begin{aligned}
s_{ij}^p = \text{cos}(\vec{w}_p \odot \vec{v}_i, \vec{w}_p \odot \vec{v}_j),\quad 
s_{ij} = \frac{1}{m}\sum_{p=1}^{m}{s_{ij}^p}
\end{aligned}
\end{equation}
Intuitively, $s_{ij}^p$ computes the cosine similarity between the two input vectors $\vec{v}_i$ and $\vec{v}_j$, for the $p$-th perspective, where each perspective considers one part of the semantics captured in the vectors.

% \subsubsection{Graph Sparsification via $\varepsilon$-neighborhood}
\textbf{{Graph sparsification via $\varepsilon$-neighborhood}.}
Typically an adjacency matrix (computed from a metric) is supposed to be non-negative but $s_{ij}$ ranges between $[-1, 1]$.
In addition, many underlying graph structures are much more sparse than a fully connected graph which is not only computationally expensive but also 
might introduce noise (i.e., unimportant edges).
% makes little sense for most applications.
We hence proceed to extract a symmetric sparse non-negative adjacency matrix $\vec{A}$ from $\vec{S}$ by considering only the $\varepsilon$-neighborhood for each node.
Specifically, we mask off (i.e., set to zero) those elements in $\vec{S}$ which are smaller than a non-negative threshold $\varepsilon$.
% \begin{equation}\label{eq:epsilon_neigh}
% % \vspace{-0.1in}
% \begin{aligned}
% A_{ij} = 
% \left\{
%         \begin{array}{ll}
%              s_{ij} & \quad  s_{ij} > \varepsilon  \\
%               0 & \quad \text{otherwise}
%         \end{array}
%     \right.
% \end{aligned}
% \end{equation}

% \subsubsection{Anchor-based Scalable Metric Learning}
\textbf{{Anchor-based scalable metric learning}.}
The above similarity metric function like~\cref{eq:similarity_metric_learning} computes similarity scores for all pairs of graph nodes, which requires $\mathcal{O}(n^2)$ complexity for both computational time and memory consumption, rendering significant scalablity issue for large graphs.
To address the scalability issue, inspired by previous anchor-based methods~\citep{liu2010large,wu2019scalable}, we design an anchor-based scalable metric learning technique which learns a node-anchor affinity matrix $\vec{R} \in \mathbb{R}^{n \times s}$ (i.e., requires $\mathcal{O}(ns)$ for both time and space complexity where $s$ is the number of anchors) between the node set $\mathcal{V}$ and the anchor set $\mathcal{U}$.
Note that $s$ is a hyperparameter which is tuned on the development set.
% instead of explicitly computing an all-to-all similarity matrix for all the nodes. 

Specifically, we randomly sample a set of $s \in \mathcal{U}$ anchors from the node set $\mathcal{V}$, where $s$ is usually much smaller than $n$ in large graphs.
The anchor embeddings are thus set to the corresponding node embeddings. 
Therefore, \cref{eq:similarity_metric_learning} can be rewritten as the following:
% \vspace{-1mm}
\begin{equation}\label{eq:anchor_metric_learning}
\begin{aligned}
a_{ik}^p = \text{cos}(\vec{w}_p \odot \vec{v}_i, \vec{w}_p \odot \vec{u}_k),\quad 
a_{ik} = \frac{1}{m}\sum_{p=1}^{m}{a_{ik}^p}
\end{aligned}
\end{equation}
where $a_{ik}$ is the affinity score between node $v_i$ and anchor $u_k$.
Similarly, we apply the $\varepsilon$-neighborhood sparsification technique to the node-anchor affinity scores $a_{ik}$ to obtain a sparse and non-negative node-anchor affinity matrix $\vec{R}$.
% \begin{equation}\label{eq:anchor_epsilon_neigh}
% % \vspace{-0.1in}
% \begin{aligned}
% R_{ik} = 
% \left\{
%         \begin{array}{ll}
%              a_{ik} & \quad  a_{ik} > \varepsilon  \\
%               0 & \quad \text{otherwise}
%         \end{array}
%     \right.
% \end{aligned}
% \end{equation}

\subsection{Graph Node Embeddings and Prediction}

% For some problems when an initial graph is available, 
% our preliminary experiments showed that it is harmful to totally discard the initial graph structure.
% Previous works~\citep{velivckovic2017graph,jiang2019semi} inject the initial graph structure into the graph learning mechanism by performing masked attention,
% which might limits its graph learning ability.
% This is because there is no way for their methods to learn weights for those
% edges that do not exist in the initial graph, but carry useful topological information.

%Once the learned graph structure $\vec{A}$ is obtained, we next have to compute the graph inputs that is best for a graph neural network. 
Although the initial graph could be noisy, it typically still carries rich and useful information regarding true graph topology. Ideally, the learned graph structure $\vec{A}$ could be supplementary to the original graph topology $\vec{A}^{(0)}$ to formulate an optimized graph for GNNs with respect to the downstream task. 
Therefore, with the mild assumption that the optimized graph structure is potentially a ``shift'' from the initial graph structure,
we combine the learned graph with the initial graph, 
\begin{equation}\label{eq:combine_adj_norm_t}
    % \small
\begin{aligned}
\widetilde{\vec{A}}^{(t)} = \lambda \vec{L}^{(0)} + (1 - \lambda) \Big\{\eta\ \text{f}(\vec{A}^{(t)}) + (1 - \eta) \ \text{f}(\vec{A}^{(1)})\Big\}
\end{aligned}
\end{equation}
% \vspace{-1mm}
where 
$\vec{L}^{(0)} = {\vec{D}^{(0)}}^{-1/2}\vec{A}^{(0)}{\vec{D}^{(0)}}^{-1/2}$
is the normalized adjacency matrix of the initial graph.
$\vec{A}^{(t)}$ and $\vec{A}^{(1)}$ are the two adjacency matrices computed at the $t$-th and 1-st iterations 
(using~\cref{eq:similarity_metric_learning}), 
% (using~\cref{eq:similarity_metric_learning,eq:epsilon_neigh}) 
% during iterative graph learning, 
respectively.
The adjacency matrix is further row normalized, namely, $\text{f}(\vec{A})_{ij} = {A_{ij}}/{\sum_j{A_{ij}}}$. 

Note that $\vec{A}^{(1)}$ is computed from the raw node features $\vec{X}$, whereas $\vec{A}^{(t)}$ is computed from the previously updated node embeddings $\vec{Z}^{(t-1)}$ that 
% usually resides on a low-dimensional manifold of the raw node feature space and 
is optimized toward the downstream prediction task. 
Therefore, we make the final learned graph structure as their linear combination weighted by a hyperparameter $\eta$, so as to combine the advantages of both. Finally, another hyperparameter $\lambda$ is used to balance the trade-off between the learned graph structure and the initial graph structure. If such an initial graph structure is not available, we instead use a kNN graph constructed based on raw node features $\vec{X}$ using cosine similarity.

Our graph learning framework is agnostic to various GNN architectures (that take as input a node feature matrix and an adjacency matrix to compute node embeddings) and prediction tasks.
In this paper,
we adopt a two-layered GCN~\citep{kipf2016semi} where the first layer (denoted as $\text{GNN}_1$) maps the raw node features $\vec{X}$ to the intermediate embedding space, and the second layer (denoted as $\text{GNN}_2$) further maps the intermediate node embeddings $\vec{Z}$ to the output space.
% \begin{align}
%   & \vec{Z} = \text{ReLU}(\widetilde{\vec{A}} \vec{X} \vec{W}_1) \label{eq:update_node_vec}\\
%   & \widehat{\vec{y}} = \sigma(\widetilde{\vec{A}} \vec{Z} \vec{W}_2) \label{eq:prediction}\\
%   & \mathcal{L}_{\text{pred}} = \ell(\widehat{\vec{y}}, \vec{y}) \label{eq:pred_loss}
% \end{align}
% \begin{align}
%   & \vec{Z} = \text{ReLU}(\text{MP}(\vec{X}, \widetilde{\vec{A}}) \vec{W}_1) \label{eq:update_node_vec}\\
%   & \widehat{\vec{y}} = \sigma(\text{MP}(\vec{Z}, \widetilde{\vec{A}}) \vec{W}_2) \label{eq:prediction}\\
%   & \mathcal{L}_{\text{pred}} = \ell(\widehat{\vec{y}}, \vec{y}) \label{eq:pred_loss}
% \end{align}
\begin{equation}
% \small
  \vec{Z} = \text{ReLU}(\text{MP}(\vec{X}, \widetilde{\vec{A}}) \vec{W}_1), \ \
  \widehat{\vec{y}} = \sigma(\text{MP}(\vec{Z}, \widetilde{\vec{A}}) \vec{W}_2), \ \
  \mathcal{L}_{\text{pred}} = \ell(\widehat{\vec{y}}, \vec{y}) \label{eq:update_node_vec-prediction-pred_loss}
\end{equation}
where 
$\sigma(\cdot)$ and $\ell(\cdot)$ are task-dependent output function and loss function, respectively.
For instance, for a classification task,
$\sigma(\cdot)$ is a softmax function for predicting a probability distribution over a set of classes, and $\ell(\cdot)$ is a cross-entropy function for computing the prediction loss.
$\text{MP}(\cdot, \cdot)$ is a message passing function, and in GCN, 
$\text{MP}(\vec{F}, \widetilde{\vec{A}})=\widetilde{\vec{A}} \vec{F}$ 
for a feature/embedding matrix $\vec{F}$ and normalized adjacency matrix $\widetilde{\vec{A}}$ which we obtain using~\cref{eq:combine_adj_norm_t}.

% \subsubsection{Node-Anchor Message Passing}
\noindent\textbf{Node-anchor message passing.}
Note that a node-anchor affinity matrix $\vec{R}$ serves as
a weighted adjacency matrix of a bipartite graph $\mathcal{B}$ allowing only direct connections between nodes and anchors.
If we regard a direct travel between a node and an anchor as one-step transition described by $\vec{R}$,
built upon theories of stationary Markov random walks~\citep{lovasz1993random},
we can actually recover both the node graph $\mathcal{G}$ and the anchor graph $\mathcal{Q}$ from $\vec{R}$ by computing the two-step transition probabilities.
Let $\vec{A}  \in \mathbb{R}^{n \times n}$ denote a row-normalized adjacency matrix for the node graph $\mathcal{G}$, 
and $A_{ij}=p^{(2)}(v_j|v_i)$ indicate the two-step transition probability from node $v_i$ to node $v_j$,
$\vec{A}$ can be recovered from $\vec{R}$,
\begin{equation}\label{eq:node_node_two_step_random_walk}
\begin{aligned}
\vec{A}= \vec{\Delta}^{-1} \vec{R} \vec{\Lambda}^{-1} \vec{R}^\top
\end{aligned}
\end{equation}
where $\Lambda \in \mathbb{R}^{s \times s}$ ($\Lambda_{kk}=\sum_{i=1}^{n}{R_{ik}}$) and $\Delta \in \mathbb{R}^{n \times n}$ ($\Delta_{ii}=\sum_{k=1}^{s}{R_{ik}}$) are both diagonal matrices. 
% where $\Lambda_{kk}=\sum_{i=1}^{n}{R_{ik}}$ and $\Delta_{ii}=\sum_{k=1}^{s}{R_{ik}}$ are $s \times s$ and $n \times n$ diagonal matrices, respectively. 
Similarly, we can recover the row-normalized adjacency matrix $\vec{B} \in \mathbb{R}^{s \times s}$ for the anchor graph $\mathcal{Q}$, 
\begin{equation}\label{eq:anchor_anchor_two_step_random_walk}
\begin{aligned}
\vec{B}= \vec{\Lambda}^{-1} \vec{R}^\top \vec{\Delta}^{-1} \vec{R}
\end{aligned}
\end{equation}
A detailed proof of recovering node and anchor graphs from the affinity matrix is provided in~\cref{sec:proof_on_two_step_random_walks}.
While explicitly computing a node adjacency matrix $\vec{A}$ from $\vec{R}$ (\cref{eq:node_node_two_step_random_walk}) and directly performing message passing over the node graph $\mathcal{G}$ (\cref{eq:update_node_vec-prediction-pred_loss}) are expensive in both time complexity ($\mathcal{O}(n^2s)$) and space complexity ($\mathcal{O}(n^2)$),
one can instead equivalently decompose the above process (denoted as $\text{MP}_{12}$) into two steps: i) node-to-anchor message passing $\text{MP}_1$ and ii) anchor-to-node message passing $\text{MP}_2$, over the node-anchor bipartite graph $\mathcal{B}$,
formulated as follows, 
\begin{equation}\label{eq:node_anchor_mp}
\begin{aligned}
\text{MP}_{12}(\vec{F}, \vec{R}) = \text{MP}_2(\vec{F}', \vec{R}), \quad \vec{F}' = \text{MP}_1(\vec{F}, \vec{R})
\end{aligned}
\end{equation}
where $\text{MP}_1(\vec{F}, \vec{R})= \vec{\Lambda}^{-1} \vec{R}^\top \vec{F}$ aims to pass message $\vec{F}$ from the nodes $\mathcal{V}$ to the anchors $\mathcal{U}$, and $\text{MP}_2(\vec{F}', \vec{R})=  \vec{\Delta}^{-1} \vec{R} \vec{F}'$ aims to further pass the message $\vec{F}'$ aggregated on the anchors back to the nodes.
Finally, we can obtain $\text{MP}_{12}(\vec{F}, \vec{R}) = \vec{\Delta}^{-1} \vec{R} \vec{\Lambda}^{-1} \vec{R}^\top \vec{F} = \vec{A} \vec{F}$ where $\vec{A}$ is the node adjacency matrix recovered from $\vec{R}$ using~\cref{eq:node_node_two_step_random_walk}.
In this way, we reduce both time and space complexity to $\mathcal{O}(ns)$.
Therefore, we can rewrite the regular node embedding and prediction equations defined in~\cref{eq:combine_adj_norm_t,eq:update_node_vec-prediction-pred_loss} as follows,
% \begin{align}
%   & \vec{Z} = \text{ReLU}(\text{MP}_{a}(\vec{X}, \{\vec{L}^{(0)}, \vec{R}^{(t)}, \vec{R}^{(1)}\} ) \vec{W}_1) \label{eq:anchor_update_node_vec}\\
%   & \widehat{\vec{y}} = \sigma(\text{MP}_{a}(\vec{Z}, \{\vec{L}^{(0)}, \vec{R}^{(t)}, \vec{R}^{(1)}\} )  \vec{W}_2) \label{eq:anchor_prediction}
% \end{align}
\begin{equation}
  \vec{Z} = \text{ReLU}(\text{MP}_{a}(\vec{X}, \{\vec{L}^{(0)}, \vec{R}^{(t)}, \vec{R}^{(1)}\} ) \vec{W}_1), \ \
  \widehat{\vec{y}} = \sigma(\text{MP}_{a}(\vec{Z}, \{\vec{L}^{(0)}, \vec{R}^{(t)}, \vec{R}^{(1)}\} )  \vec{W}_2) \label{eq:anchor_update_node_vec-prediction}
\end{equation}
where $\text{MP}_a(\cdot, \cdot)$ is a hybrid message passing function with the same spirit of~\cref{eq:combine_adj_norm_t}, defined as,
\begin{align}
\mbox{\footnotesize\( %
\text{MP}_{a}(\vec{F}, \{\vec{L}^{(0)}, \vec{R}^{(t)}, \vec{R}^{(1)}\} ) = \lambda \text{MP}(\vec{F}, \vec{L}^{(0)}) + (1-\lambda) \Big\{ \eta \text{MP}_{12}(\vec{F}, \vec{R}^{(t)}) + (1-\eta) \text{MP}_{12}(\vec{F}, \vec{R}^{(1)}) \Big\}
\)} %
\label{eq:anchor_combine_adj_norm_t}
\end{align}
Note that we use the same $\text{MP}(\cdot, \cdot)$ function defined in~\cref{eq:update_node_vec-prediction-pred_loss} for performing message passing over $\vec{L}^{(0)}$ which is typically sparse in practice, 
and $\vec{F}$ can either be $\vec{X}$ or $\vec{Z}$.

% \subsection{Graph learning from smooth signals}
\subsection{Graph Regularization}
Although combining the learned graph $\vec{A}^{(t)}$ with the initial graph $\vec{A}^{(0)}$ is an effective way to approach the optimaized graph, the quality of the learned graph $\vec{A}^{(t)}$ plays an important role in improving the quality of the final graph $\widetilde{\vec{A}}^{(t)}$. In practice, it is important to control the smoothness, connectivity and sparsity of the resulting learned graph $\vec{A}^{(t)}$, which faithfully reflects the graph topology with respect to the initial node attributes $\vec{X}$ and the downstream task. 

% In graph signal processing, various methods~\citep{shuman2013emerging,kalofolias2016learn,kalofolias2017large} have been explored to construct a graph from data points.
% These methods usually solve an optimization problem with certain assumptions on the graph signals or structural constraints on the underlying graphs, without considering a downstream task. 
% Inspired by these techniques, we adapt them and apply them as regularization terms to the learned graph $\vec{A}^{(t)}$.

Let each column of the feature matrix $\vec{X}$ be considered as a graph signal. A widely adopted assumption for graph signals is that values change smoothly across adjacent nodes.
Given an undirected graph with a symmetric weighted adjacency matrix $A$, the smoothness of a set of $n$ graph signals $\vec{x}_1, \dots, \vec{x}_n \in \mathbb{R}^{d}$ is usually measured by the Dirichlet energy~\citep{belkin2002laplacian}, 
\begin{equation}\label{eq:smoothness_loss}
\begin{aligned}
\Omega(\vec{A}, \vec{X})=\frac{1}{2 n^2}\sum_{i,j}A_{ij}||\vec{x}_i - \vec{x}_j||^2=\frac{1}{n^2}\text{tr}(\vec{X}^T \vec{L} \vec{X})
\end{aligned}
\end{equation}
where $\text{tr}(\cdot)$ denotes the trace of a matrix,
$\vec{L}=\vec{D}-\vec{A}$ is the graph Laplacian, and $\vec{D}=\sum_{j}A_{ij}$ is the degree matrix. 
As can be seen, minimizing $\Omega(\vec{A}, \vec{X})$ forces adjacent nodes to have similar features, thus enforcing
smoothness of the graph signals on the graph associated with $\vec{A}$.

% %%%%%%%%%%%%%%%%%%%
% Unified algorithm for IDGL and IDGL-anchor - NeurIPS
% %%%%%%%%%%%%%%%%%%%
\begin{algorithm}[!tbh]
\small
%  \algsetup{linenosize=\tiny}
% \caption{Iterative Deep Graph Learning Framework}
\caption{General Framework for \alg and \salg}
\label{alg:IDGL}
\begin{algorithmic}[1]
\STATE {\bfseries Input:} $\vec{X}$, $\vec{y}$$[, \vec{A}^{(0)}]$
\STATE {\bfseries Parameters:} $m$, $\varepsilon$, $\alpha$, $\beta$, $\gamma$, $\lambda$, $\delta$, $T$, $\eta$, $k$$[, s]$
\STATE {\bfseries Output:} $\Theta$, $\widehat{\vec{y}}$, $\widetilde{\vec{A}}^{(t)}$ or $\vec{R}^{(t)}$

\STATE $[\vec{A}^{(0)} \leftarrow  \text{kNN}( \vec{X}, k)]$ 
\COMMENT{kNN-graph if no initial $\vec{A}^{(0)}$}
 
\STATE $t \leftarrow 1$\;

% Stopping condition
% \COMMENT{Stopping condition}
\STATE $\text{StopCond} \leftarrow |\vec{A}^{(t)} - \vec{A}^{(t-1)}|_F^2 > \delta |\vec{A}^{(1)}|_F^2$ \ \textbf{if} \  \alg \  \textbf{else} \  $|\vec{R}^{(t)} - \vec{R}^{(t-1)}|_F^2 > \delta |\vec{R}^{(t)}|_F^2$

\WHILE{($(t == 1 \ \OR \ \text{StopCond}$ )  \AND $t \leq T$}\label{alg_line:stopping_criterion}
    
    \IF{\alg}    

        \STATE    $\vec{A}^{(t)} \leftarrow \text{GL}(\vec{X}) \ \OR \ \text{GL}(\vec{Z}^{(t-1)})$ 
        % using~\cref{eq:similarity_metric_learning,eq:epsilon_neigh}
        using~\cref{eq:similarity_metric_learning}
        \label{alg_line:refine_adj}
            \COMMENT{Refine adj. matrix}

        \STATE    $\widetilde{\vec{A}}^{(t)} \leftarrow 
        % \{\vec{A}^{(0)}, \vec{A}^{(t)} \} \ \OR \ \{\vec{A}^{(0)},
         \{\vec{A}^{(0)},
        \vec{A}^{(t)}, \vec{A}^{(1)}\}$ using~\cref{eq:combine_adj_norm_t}
            \label{alg_line:combine_adj_norm_t}
            \COMMENT{Combine refined and raw adj. matrices}

        \STATE    $\vec{Z}^{(t)} \leftarrow \text{GNN}_1(\widetilde{\vec{A}}^{(t)}, \vec{X})$ using~\cref{eq:update_node_vec-prediction-pred_loss}
            \label{alg_line:refine_node_vec}
            \COMMENT{Refine node embeddings}

        % \STATE        $\widehat{\vec{y}} \leftarrow \text{GNN}_2(\widetilde{\vec{A}}^{(t)}, \vec{Z}^{(t)})$ using~\cref{eq:prediction}
        %         % \COMMENT{Compute task output}

    \ELSE
        
        \STATE    $\vec{R}^{(t)} \leftarrow \text{GL}(\vec{X}, \vec{X}_\mathcal{U}) \ \OR \ \text{GL}(\vec{Z}^{(t-1)}, \vec{Z}^{(t-1)}_\mathcal{U})$ 
        % using~\cref{eq:anchor_metric_learning,eq:anchor_epsilon_neigh}
        using~\cref{eq:anchor_metric_learning}
        \label{alg_line:anchor_refine_adj}
            \COMMENT{Refine affinity matrix}

        \STATE    $\vec{Z}^{(t)} \leftarrow \text{GNN}_1(\{\vec{A}^{(0)}, \vec{R}^{(t)}, \vec{R}^{(1)}\}, \vec{X})$ using~\cref{eq:anchor_combine_adj_norm_t,eq:anchor_update_node_vec-prediction}
            \label{alg_line:anchor_refine_node_vec}
            \COMMENT{Refine node embeddings}

        % \STATE        $\widehat{\vec{y}} \leftarrow \text{GNN}_2(\{\vec{A}^{(0)}, \vec{R}^{(t)}, \vec{R}^{(1)}\}, \vec{Z}^{(t)})$ using~\cref{eq:anchor_combine_adj_norm_t,eq:anchor_prediction}
        %         % \COMMENT{Compute task output}

    \ENDIF

    % \STATE    $\vec{Z}^{(t)} \leftarrow \text{GNN}_1(\widetilde{\vec{A}}^{(t)}, \vec{X})$ using~\cref{eq:update_node_vec} \ \textbf{if} \ \alg \ \textbf{else} \ 
    % $\text{GNN}_1(\{\vec{A}^{(0)}, \vec{R}^{(t)}, \vec{R}^{(1)}\}, \vec{X})$
    % using~\cref{eq:anchor_combine_adj_norm_t,eq:anchor_update_node_vec}\;
    % \label{alg_line:refine_node_vec}
    % \COMMENT{Refine node embeddings}

    \STATE        $\widehat{\vec{y}} \leftarrow \text{GNN}_2(\widetilde{\vec{A}}^{(t)}, \vec{Z}^{(t)})$ using~\cref{eq:update_node_vec-prediction-pred_loss} \ \textbf{if} \ \alg \ \textbf{else} \ 
    $\text{GNN}_2(\{\vec{A}^{(0)}, \vec{R}^{(t)}, \vec{R}^{(1)}\}, \vec{Z}^{(t)})$ using~\cref{eq:anchor_combine_adj_norm_t,eq:anchor_update_node_vec-prediction}

    \STATE        $\mathcal{L}_{\text{pred}}^{(t)} \leftarrow \text{LOSS}_1(\widehat{\vec{y}}, \vec{y})$ using~\cref{eq:update_node_vec-prediction-pred_loss}\;

    % \STATE        $\mathcal{L}_{\mathcal{G}}^{(t)} \leftarrow
    % \text{LOSS}_2(\vec{A}^{(t)}, \vec{X})$ 
    % using~\cref{eq:graph_loss} \ \textbf{if} \ \alg \ \textbf{else} \ 
    % $\text{LOSS}_2(\vec{R}^{(t)}, \vec{X}_\mathcal{U})$ using~\cref{eq:anchor_graph_loss}\; \label{alg_line:graph regularization}

    \STATE        $\mathcal{L}_{\mathcal{G}}^{(t)} \leftarrow
    \alpha   \Omega(\vec{A}^{(t)}, \vec{X}) + f(\vec{A}^{(t)})$ 
    \ \textbf{if} \ \alg \ \textbf{else} \ 
    $\alpha   \Omega(\widehat{\vec{B}}^{(t)}, \vec{X}^{\mathcal{U}}) + f(\widehat{\vec{B}}^{(t)})$ \  \textbf{where} \ 
    $\widehat{\vec{B}}^{(t)}= {\vec{R}^{(t)}}^\top \vec{\Delta}^{-1} \vec{R}^{(t)}$
    \; \label{alg_line:graph regularization}

    \STATE        $\mathcal{L}^{(t)} \leftarrow  \mathcal{L}_{\text{pred}}^{(t)} + \mathcal{L}_{\mathcal{G}}^{(t)}$\;\label{alg_line:joint_loss_t}\ \AND \ $t \leftarrow t + 1$\;

\ENDWHILE

\STATE    $\mathcal{L} \leftarrow \mathcal{L}^{(1)}  + \sum_{i=2}^t{\mathcal{L}^{(i)}} / (t - 1)$\;

% \IF{Training}
%     \STATE Back-propagate $\mathcal{L}$ to update model weights $\Theta$\;\label{alg_line:BP}
% \ENDIF

 \STATE Back-propagate $\mathcal{L}$ to update model weights $\Theta$\;\label{alg_line:BP}
  \COMMENT{In training phase only}

\end{algorithmic}
\end{algorithm}

However, solely minimizing the smoothness loss will result in the trivial solution $\vec{A}=0$.
Also, it is desirable to have control of how sparse the resulting graph is.
Following~\citep{kalofolias2016learn},
we impose additional constraints on the learned graph,
\begin{equation}\label{eq:degree_sparsity_loss}
\begin{aligned}
f(\vec{A})= \frac{-\beta}{n}\vec{1}^T \text{log}(\vec{A}\vec{1}) +  \frac{\gamma}{n^2} ||\vec{A}||_F^2
\end{aligned}
\end{equation}
where $||\cdot||_F$ denotes the Frobenius norm of a matrix.
The first term penalizes the formation of disconnected graphs via the logarithmic barrier,
and the second term controls sparsity by penalizing large degrees due to the first term.

We then define the overall graph regularization loss as the sum of the above losses $\mathcal{L}_{\mathcal{G}} = \alpha   \Omega(\vec{A}, \vec{X}) + f(\vec{A})$, which is able to control the smoothness, connectivity and sparsity of the learned graph where $\alpha$, $\beta$ and $\gamma$ are all non-negative hyperparameters.
% \begin{equation}\label{eq:graph_loss}
% \begin{aligned}
% \mathcal{L}_{\mathcal{G}} = \alpha   \Omega(\vec{A}, \vec{X}) + f(\vec{A}).
% \end{aligned}
% \end{equation}

\noindent\textbf{Anchor graph regularization.}
As shown in~\cref{eq:anchor_anchor_two_step_random_walk}, we can obtain a row-normalized adjacency matrix $\vec{B}$ for the anchor graph $\mathcal{Q}$ in $\mathcal{O}(ns^2)$ time complexity.
In order to control the quality of the learned node-anchor affinity matrix $\vec{R}$ (which can result in implicit control of the quality of the node adjacency matrix $\vec{A}$), we apply the aforementioned graph regularization techniques to the anchor graph.
It is worthing noting that our proposed graph regularization loss is only applicable to non-negative and symmetric adjacency matrices~\citep{kalofolias2017large}.
Therefore, instead of applying graph regularization to $\vec{B}$ which is often not symmetric,
we opt to apply graph regularization to its unnormalized version $\widehat{\vec{B}}= \vec{R}^\top \vec{\Delta}^{-1} \vec{R}$ as $\mathcal{L}_{\mathcal{G}} = \alpha   \Omega(\widehat{\vec{B}}, \vec{X}^{\mathcal{U}}) + f(\widehat{\vec{B}})$,
% \begin{equation}\label{eq:anchor_graph_loss}
% \begin{aligned}
% \mathcal{L}_{\mathcal{G}} = \alpha   \Omega(\widehat{\vec{B}}, \vec{X}^{\mathcal{U}}) + f(\widehat{\vec{B}}).
% \end{aligned}
% \end{equation}
where $\vec{X}^{\mathcal{U}}$ denotes the set of anchor embeddings sampled from the set of node embeddings $\vec{X}$.

\subsection{Joint Learning with A Hybrid Loss}

Compared to previous works which directly optimize the adjacency matrix based on either graph regularization loss~\citep{kalofolias2017large}, or task-dependent prediction loss~\citep{franceschi2019learning},
we propose to jointly and iteratively learning the graph structure and the GNN parameters by 
% we propose to learn an optimized graph structure through a similarity metric function and the GNN parameters by 
minimizing a hybrid loss function combining both the task prediction loss and the graph regularization loss, namely, $\mathcal{L} = \mathcal{L}_{\text{pred}} + \mathcal{L}_{\mathcal{G}}$.
% One big advantage of learning a similarity metric function instead of an adjacency matrix is that the learned model can be easily generalized to unseen graphs in an inductive setting.

The full algorithm of the \alg framework is presented in~\cref{alg:IDGL}. 
As we can see, our model repeatedly refines the adjacency matrix with updated node embeddings 
% (\cref{eq:similarity_metric_learning,eq:epsilon_neigh}),
(\cref{eq:similarity_metric_learning}),
and refines the node embeddings (\cref{eq:combine_adj_norm_t,eq:update_node_vec-prediction-pred_loss}) with the updated adjacency matrix
% (Line \ref{alg_line:refine_adj}-\ref{alg_line:refine_node_vec}),
% until some stopping criterion (Line~\ref{alg_line:stopping_criterion}) is satisfied.
until the difference between adjacency matrices at consecutive iterations are smaller than certain threshold.
Note that compared to using a fixed number of iterations globally, 
our dynamic stopping criterion is more beneficial, especially for mini-batch training.
 At each iteration, a hybrid loss combining both the task-dependent prediction loss and the graph regularization loss is computed.
%  (Line~\ref{alg_line:joint_loss_t}).
After all iterations,
the overall loss is back-propagated through all previous iterations to update the model parameters.
% (Line~\ref{alg_line:BP}).
Notably, \cref{alg:IDGL} is also applicable to \salg. The major differences between \alg and \salg are how we compute adjacency (or affinity) matrix, and perform message passing and graph regularization.

\section{Experiments}

In this section, we conduct extensive experiments to verify the effectiveness of \alg and \salg in various settings. 
The implementation of our proposed models is publicly available at \url{https://github.com/hugochan/IDGL}.
% The code and data for our proposed models are provided for research purpose (Code is included with the submission and will be released upon the paper acceptance).
% \footnotemark[1].
% \footnotetext[1]{Code is included with the submission. Code will be released upon the paper acceptance.}
% The details on model settings are provided in~\cref{sec:model_settings}.
% Our code and data will be made publicly available upon acceptance of this paper.

% \subsection{Datasets and Baselines}
\textbf{{Datasets and baselines.}}
The benchmarks used in our experiments include
four citation network datasets (i.e., Cora, Citeseer, Pubmed and ogbn-arxiv)~\citep{sen2008collective,hu2020open} where the graph topology is available,
three non-graph datasets (i.e., Wine, Breast Cancer (Cancer) and Digits)~\citep{Dua:2019} where the graph topology does not exist,
and two text benchmarks (i.e., 20Newsgroups data (20News) and movie review data (MRD))~\citep{lang1995newsweeder,pang2004sentimental}
where we treat a document as a graph containing each word as a node.
The first seven datasets are all for node classification tasks in the transductive setting, and we follow the experimental setup of previous works~\citep{kipf2016semi,franceschi2019learning,hu2020open}.
The later two datasets are for graph-level prediction tasks in the inductive setting.
Please refer to~\cref{sec:data_statistics} for detailed data statistics.

% The benchmarks used in our experiments include 
% four graph benchmarks, three data point benchmarks and two text benchmarks.
% Our graph benchmarks include Cora, Citeseer, Pubmed and ogbn-arxiv citation network datasets, which are commonly used for evaluating graph-based learning algorithms~\citep{sen2008collective,hu2020open}.
% In addition to graph benchmarks where the graph topology is available,
% we include three data point benchmarks (i.e., Wine, Breast Cancer (Cancer) and Digits from the UCI machine learning repository~\citep{Dua:2019}).
% The above seven datasets are all for node classification tasks, and we follow the experimental setup of previous works~\citep{kipf2016semi,franceschi2019learning,hu2020open}.
% Finally, to demonstrate the effectiveness of our models on inductive learning problems, we conduct document classification and regression tasks on the 20Newsgroups data (20News)~\citep{lang1995newsweeder} and the movie review data (MRD)~\citep{pang2004sentimental}, respectively, where we treat a document as a graph containing each word as a node.
% % For 20News, we randomly select 30\% examples from the training data as the development set.
% % For MRD, we split the data to train/dev/test sets using a 60\%/20\%/20\% split.
% Please refer to~\cref{sec:data_statistics} for detailed data statistics.

Our main baseline is LDS~\citep{franceschi2019learning} which however is incapable of handling inductive learning problems, we hence only report its results on transductive datasets.
% The experimental results of several semi-supervised (i.e., SemiEmb~\citep{weston2012deep}) and supervised learning (i.e., RBF SVM) baselines are taken from the LDS paper.
% For the sake of completeness, we directly copy their results here.
% For ease of comparison, we also copy the reported results of LDS even though we rerun the experiments of LDS using the official code released by the authors.
In addition, for citation network datasets, we include other GNN variants (i.e., GCN~\citep{kipf2016semi}, GAT~\citep{velivckovic2017graph}, GraphSAGE~\citep{hamilton2017inductive}, APPNP~\citep{klicpera2018predict}, H-GCN~\citep{hu2019semi} and GDC~\citep{klicpera2019diffusion}) as baselines.
% The GraphSage results are taken from ~\citep{shchur2018pitfalls}.
For non-graph and text benchmarks where the graph topology is unavailable,
we conceive various GNN$_{\text{kNN}}$ baselines (i.e., GCN$_{\text{kNN}}$, GAT$_{\text{kNN}}$ and GraphSAGE$_{\text{kNN}}$) where a kNN graph on the data set is constructed during preprocessing before applying a GNN model.
For text benchmarks, we include a BiLSTM~\citep{hochreiter1997long} baseline.
The reported results are averaged over 5 runs with different random seeds.

% \subsection{Experimental Results}
\textbf{{Experimental results.}}
% \cref{table:transductive_results} shows the results of transductive experiments.
% First of all, we can see that \alg outperforms all baselines in 4 out of 5 benchmarks, which demonstrates its effectiveness.
% Compared to \alg, \salg is more scalable and can achieve comparable or even better results.
% Besides, 
% % by comparing the results of GCN, GAT and \alg on Cora and Citeseer, and considering that GCN is the underlying GNN module of \alg, 
% we can see that our graph learning method can greatly help the node classification task even when the graph topology is given.
% When the graph topology is not available, 
% compared to various GNN$_{\text{kNN}}$ baselines, \alg consistently achieves much better results on all datasets, which shows the power of jointly learning graph structures and GNN parameters.
% The results of inductive experiments are shown in~\cref{table:inductive_results}. 
% Unlike LDS which cannot handle inductive setting, 
% the good performance on 20News and MRD verifies the capability of \alg on inductive learning.
%%%%%%%%%%%%%%%%%%%%%%%%%%%%%%%%%%%%%%%%%%
\cref{table:transductive_results} shows the results of transductive experiments.
First of all, \alg outperforms all baselines in 4 out of 5 benchmarks.
Besides, compared to \alg, \salg is more scalable and can achieve comparable or even better results.
In a scenario where the graph structure is available, compared to the state-of-the-art GNNs and graph learning models, our models achieve either significantly better or competitive results, even though the underlying GNN component of our models is a vanilla GCN.
When the graph topology is not available (thus GNNs are not directly applicable), 
compared to graph learning baselines, \alg consistently achieves much better results on all datasets.
Compared to our main graph learning baseline LDS, our models not only achieve significantly better performance, but also are more scalable.
The results of inductive experiments are shown in~\cref{table:inductive_results}. 
Unlike LDS which cannot handle inductive setting, 
the good performance on 20News and MRD demonstrates the capability of \alg on inductive learning.

\begin{table}[!tbh]
% \vspace{-2mm}
\caption{
% Test accuracy ($\pm$ standard deviation) in percent on various classification datasets in the transductive setting. 
Summary of results in terms of classification accuracies (in percent) on transductive benchmarks.
The star symbol indicates that we ran the experiments. The dash symbol indicates that reported results were unavailable or we were not able to run the experiments due to memory issue.
}
\label{table:transductive_results}
\centering
\addtolength{\tabcolsep}{-1pt}
\scalebox{0.89}{
\begin{tabular}{llllllllll}
\hline
  Model & \vline & Cora & Citeseer & Pubmed & ogbn-arxiv& Wine & Cancer & Digits\\
  \hline
%   LogReg & \vline & 60.8 (0.0)&62.2 (0.0)& \quad\textrm{---}& \quad\textrm{---}& 92.1 (1.3)&93.3 (0.5)&85.5 (1.5)\\
%   Linear SVM & \vline & 58.9 (0.0)&58.3 (0.0)&93.9 (1.6) &90.6 (4.5)& 87.1 (1.8)\\
%   RBF SVM & \vline &59.7 (0.0) &60.2 (0.0)& \quad\textrm{---}& \quad\textrm{---}&94.1 (2.9) &91.7 (3.1)&86.9 (3.2)\\
%   RF & \vline & 58.7 (0.4)&60.7 (0.7)& 93.7 (1.6)&92.1 (1.7)& 83.1 (2.6)\\
%   FFNN & \vline & 56.1 (1.6)&56.7 (1.7)& \quad\textrm{---}& \quad\textrm{---}&89.7 (1.9) &92.9 (1.2)&36.3 (10.3)\\
%   LP & \vline &37.8 (0.2) &23.2 (6.7)& \quad\textrm{---}& \quad\textrm{---}&89.8 (3.7) &76.6 (0.5)& 91.9 (3.1)\\
%   ManiReg & \vline &62.3 (0.9) &67.7 (1.6)& \quad\textrm{---}& \quad\textrm{---}& 90.5 (0.1)&81.8 (0.1)&83.9 (0.1)\\
%   SemiEmb & \vline &63.1 (0.1) &68.1 (0.1)&\quad\textrm{---}& \quad\textrm{---} &91.9 (0.1) &89.7 (0.1)& 90.9 (0.1)\\
  GCN & \vline & 81.5&70.3&79.0& 71.7 (0.3)&\quad\textrm{---} &\quad\textrm{---}&\quad\textrm{---}\\
  GAT & \vline &83.0 (0.7) &72.5 (0.7)&79.0 (0.3) &\quad\textrm{---}& \quad\textrm{---} &\quad\textrm{---}&\quad\textrm{---}\\
GraphSAGE & \vline & 77.4 (1.0) & 67.0 (1.0) & 76.6 (0.8) &71.5 (0.3)& \quad\textrm{---} &\quad\textrm{---}&\quad\textrm{---}\\
APPNP & \vline & \quad\textrm{---} &  \textbf{75.7 (0.3)} & 79.7 (0.3)  & \quad\textrm{---} & \quad\textrm{---} &\quad\textrm{---}&\quad\textrm{---}\\
H-GCN & \vline & \textbf{84.5 (0.5)}  &  72.8 (0.5) & 79.8 (0.4)  & \quad\textrm{---} & \quad\textrm{---}  & \quad\textrm{---} & \quad\textrm{---} \\
GCN+GDC & \vline & 83.6 (0.2)  &  73.4 (0.3)  &  78.7 (0.4) & \quad\textrm{---} & \quad\textrm{---}  & \quad\textrm{---} & \quad\textrm{---} \\
  LDS & \vline & 84.1 (0.4)&75.0 (0.4)& \quad\textrm{---}& \quad\textrm{---}& 97.3 (0.4) & 94.4 (1.9) & 92.5 (0.7)\\
  \hline
%   GCN* & \vline & 81.0 (0.2)&70.9 (0.3)&79.2 (0.1)& 71.4 (0.2) &\quad\textrm{---} & \quad\textrm{---}&\quad\textrm{---}\\
  GCN$_{\text{kNN}}*$ & \vline & \quad\textrm{---} &\quad\textrm{---}& \quad\textrm{---} & \quad\textrm{---}&95.9 (0.9) &94.7 (1.2) &89.5 (1.3) \\
  GAT$_{\text{kNN}}*$ & \vline & \quad\textrm{---} &\quad\textrm{---}& \quad\textrm{---} & \quad\textrm{---}&95.8 (3.1) & 88.6 (2.7) & 89.8 (0.6) \\
  GraphSAGE$_{\text{kNN}}*$ & \vline & \quad\textrm{---} &\quad\textrm{---}& \quad\textrm{---} & \quad\textrm{---}&96.5 (1.1)  &92.8 (1.0) & 88.4 (1.8) \\
%   GAT* & \vline &82.5 (0.4) &70.9 (0.4)&\quad\textrm{---}& \quad\textrm{---} &\quad\textrm{---} &\quad\textrm{---}&\quad\textrm{---}\\
  LDS* & \vline & 83.9 (0.6)& 74.8 (0.3)& \quad\textrm{---}& \quad\textrm{---}& 96.9 (1.4) &93.4 (2.4)&90.8 (2.5)\\
    \hline
  \alg & \vline & \textbf{84.5 (0.3)}&74.1 (0.2)&\quad\textrm{---}& \quad\textrm{---} &97.8 (0.6) & \textbf{95.1 (1.0)}& 93.1 (0.5)\\
    
    \salg & \vline & 84.4 (0.2) & 72.0 (1.0)& \textbf{83.0 (0.2)} & \textbf{72.0 (0.3)} & \textbf{98.1 (1.1)} & 94.8 (1.4) & \textbf{93.2 (0.9)}\\
 \hline
\end{tabular}
}
% \vspace{-2mm}
\end{table}

% Inductive setting
\begin{table}[tbh]
% \begin{wraptable}{r}{2.6in}
% \vspace{-4mm}
\caption{
Summary of results in terms of classification accuracies or regression scores ($R^2$) (in percent) on inductive benchmarks.
% Test scores in percent on classification (accuracy) and regression ($R^2$) datasets in the inductive setting.
}
\label{table:inductive_results}
\centering
%  \small
% \addtolength{\tabcolsep}{-4pt}
\scalebox{1}{
\begin{tabular}{llll}
\hline
  Methods & \vline & 20News & MRD \\
    \hline
  BiLSTM & \vline & 80.0 (0.4) &53.1 (1.4)\\
  GCN$_{\text{kNN}}$ & \vline & 81.3 (0.6)&60.1 (1.5)\\
      \hline
  \alg & \vline & \textbf{83.6 (0.4)} & \textbf{63.7 (1.8)}\\
  \salg & \vline &82.9 (0.3) & 62.9 (0.4)\\
 \hline
\end{tabular}
}
% \vspace{-4mm}
\end{table}
% \end{wraptable}

\begin{table}[!tbh]
% \vspace{-4mm}
\caption{Ablation study on various node/graph classification datasets. 
% (reg. indicates regularization and IL indicates iterative learning.)
} 
\label{table:ablation_results}
\centering
\scalebox{1}{
\begin{tabular}{llllllllll}
\hline
  Methods & \vline & Cora & Citeseer & Wine & Cancer & Digits & 20News\\
  \hline
\alg & \vline & \textbf{84.5 (0.3)} & \textbf{74.1 (0.2)} & \textbf{97.8 (0.6)} & \textbf{95.1 (1.0)} & \textbf{93.1 (0.5)} & \textbf{83.6 (0.4)}\\
w/o graph reg. & \vline & 84.3 (0.4)&71.5 (0.9)&97.3 (0.8) &94.9 (1.0) & 91.5 (0.9)& 83.4 (0.5)\\
w/o IL & \vline &83.5 (0.6) &71.0 (0.8)&97.2 (0.8) &94.7 (0.9)&92.4 (0.4) & 83.0 (0.4)\\
  \hline
\salg & \vline & \textbf{84.4 (0.2)} & \textbf{72.0 (1.0)} & \textbf{98.1 (1.1)} & \textbf{94.8 (1.4)} & \textbf{93.2 (0.9)} & \textbf{82.9 (0.3)}\\
w/o graph reg. & \vline &83.2 (0.8) & 70.1 (0.8) & 97.4 (1.8)& 
\textbf{94.8 (1.4)}& 92.0 (1.3) &82.5 (0.7)\\
w/o IL & \vline &  83.6 (0.2)&68.6 (0.7) & 96.4 (1.5)&
94.0 (2.6)&93.0 (0.4) &82.3 (0.3)\\
 \hline
\end{tabular}
}
% \vspace{-4mm}
\end{table}

% \subsection{Ablation Study}
\textbf{{Ablation study.}}
% We perform an ablation study to assess the impact of different modules in our models.
% As shown in~\cref{table:ablation_results}, 
~\cref{table:ablation_results} shows the ablation study results on different modules in our models.
we can see a significant performance drop consistently for both \alg and \salg on all datasets by turning off the iterative learning component (i.e., iterating only once), indicating its effectiveness.
Besides, we can see the benefits of jointly training the model with the graph regularization loss.
% For instance, when training the model without the graph regularization loss, the performance on Citeseer drops from 74.1\% to 71.5\%.

\begin{figure}[th]
%   \vspace{-0.1in}
  \centerline{
  \subfloat[t][Edge deletion]{%
    \includegraphics[keepaspectratio=true,scale=0.14]{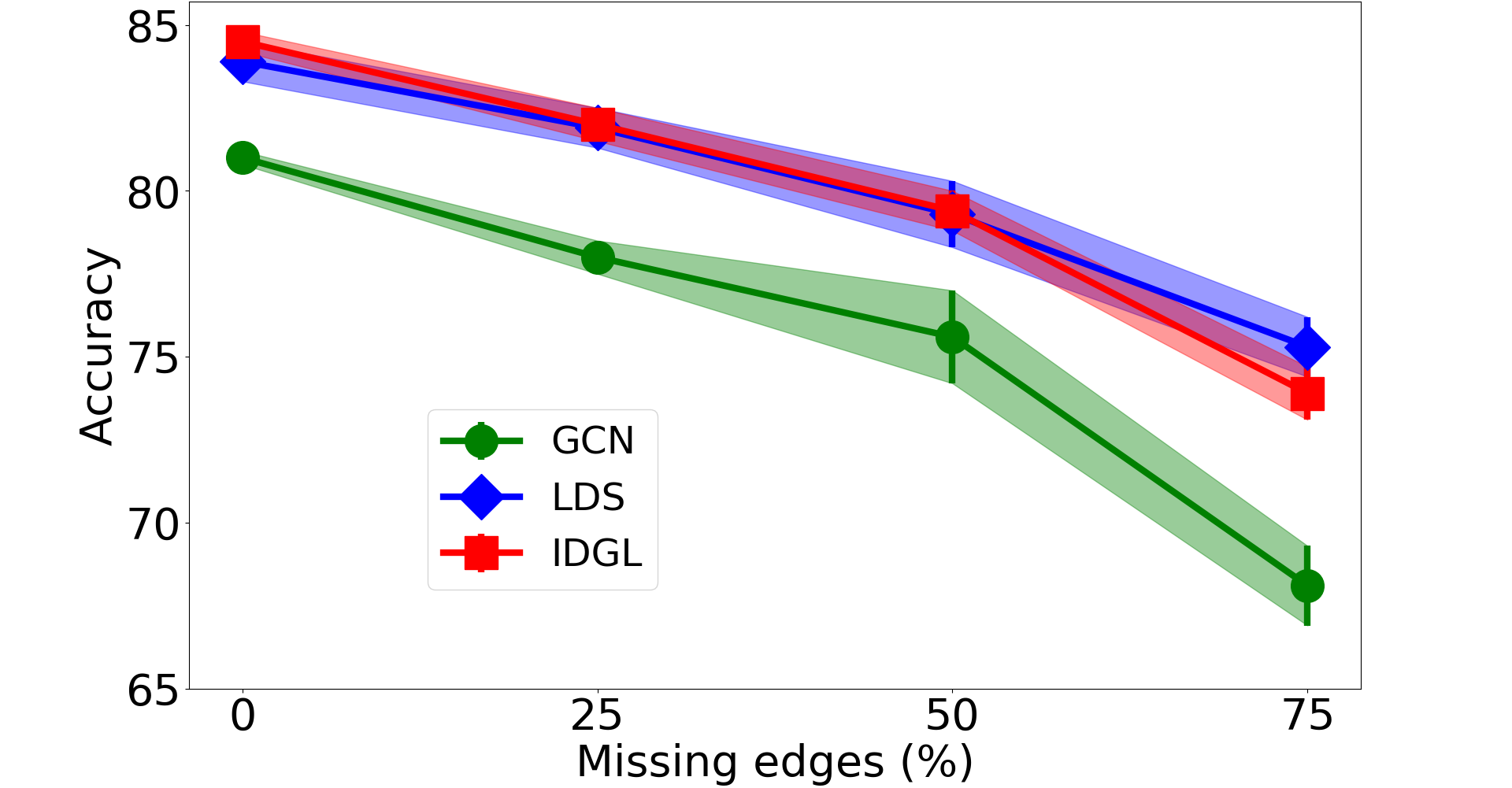}
	\label{fig:cora_missing_edges}
	}
% \hspace{0.1in}
  \subfloat[t][Edge addition]{%
    \includegraphics[keepaspectratio=true,scale=0.14]{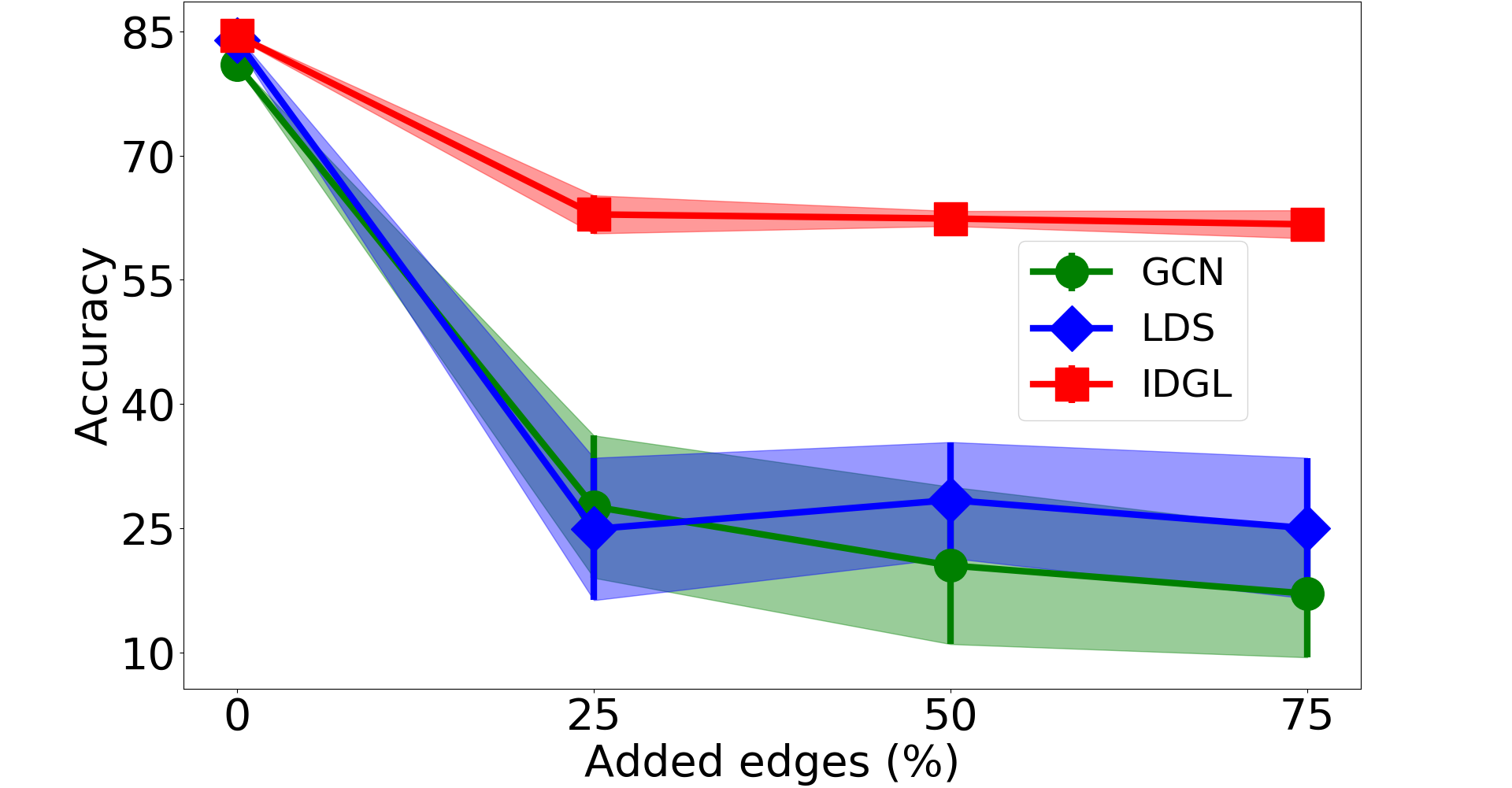}
	\label{fig:cora_added_edges}
	}
 }
%   \vspace{-0.1in}
  \caption{Test accuracy ($\pm$ standard deviation) in percent for the edge attack scenarios on Cora.}
  \label{fig:cora_edge_attack_plot}
%   \vspace{-0.1in}
\end{figure}

% \subsection{Model Analysis}
\textbf{{Model analysis.}}
% In this subsection, we analyze the proposed model in multiple dimensions. 
% First, we evaluate the robustness of \alg to adversarial graphs with random edge deletions or additions. Second, we empirically examine whether the iterative learning procedure can converge. Third, we explore and compare different stopping strategies for the iterative learning method. Finally, we empirically compare \alg with other baseline methods in terms of training efficiency.
% \noindent\textbf{Robustness analysis.}
To evaluate the robustness of \alg to adversarial graphs, we construct graphs with random edge deletions or additions.
% Specifically, we randomly remove or add 25\%, 50\% and 75\% of the edges in the original graphs.
Specifically, for each pair of nodes in the original graph, we randomly remove (if an edge exists) or add (if no such edge) an edge with a probability 25\%, 50\% or 75\%.
As shown in~\cref{fig:cora_edge_attack_plot}, compared to GCN and LDS, \alg achieves better or comparable results in both scenarios.
% Notably, the performance drop of \alg is slower than that of GCN when increasing the percent of missing edges.
% Compared to GCN, the accuracy gain of \alg changes from 3.5\% to 5.6\% on Cora, and from 3.2\% to 4.9\% on Citeseer when the missing ratio increases.
While both GCN and LDS completely fail in the edge addition scenario, \alg performs reasonably well. 
We conjecture this is because
the edge addition scenario is more challenging than the edge deletion scenario by incorporating misleading additive random noise to the initial graph. 
And \cref{eq:combine_adj_norm_t} is formulated as a form of skip-connection, by lowering the value of $\lambda$ (i.e., tuned on the development set), we enforce the model to rely less on the initial noisy graph.

\begin{figure}[!th]
%   \vspace{-0.3in}
  \centerline{
  \subfloat[t][Convergence plot on Cora.]{%
    \includegraphics[keepaspectratio=true,scale=0.14]{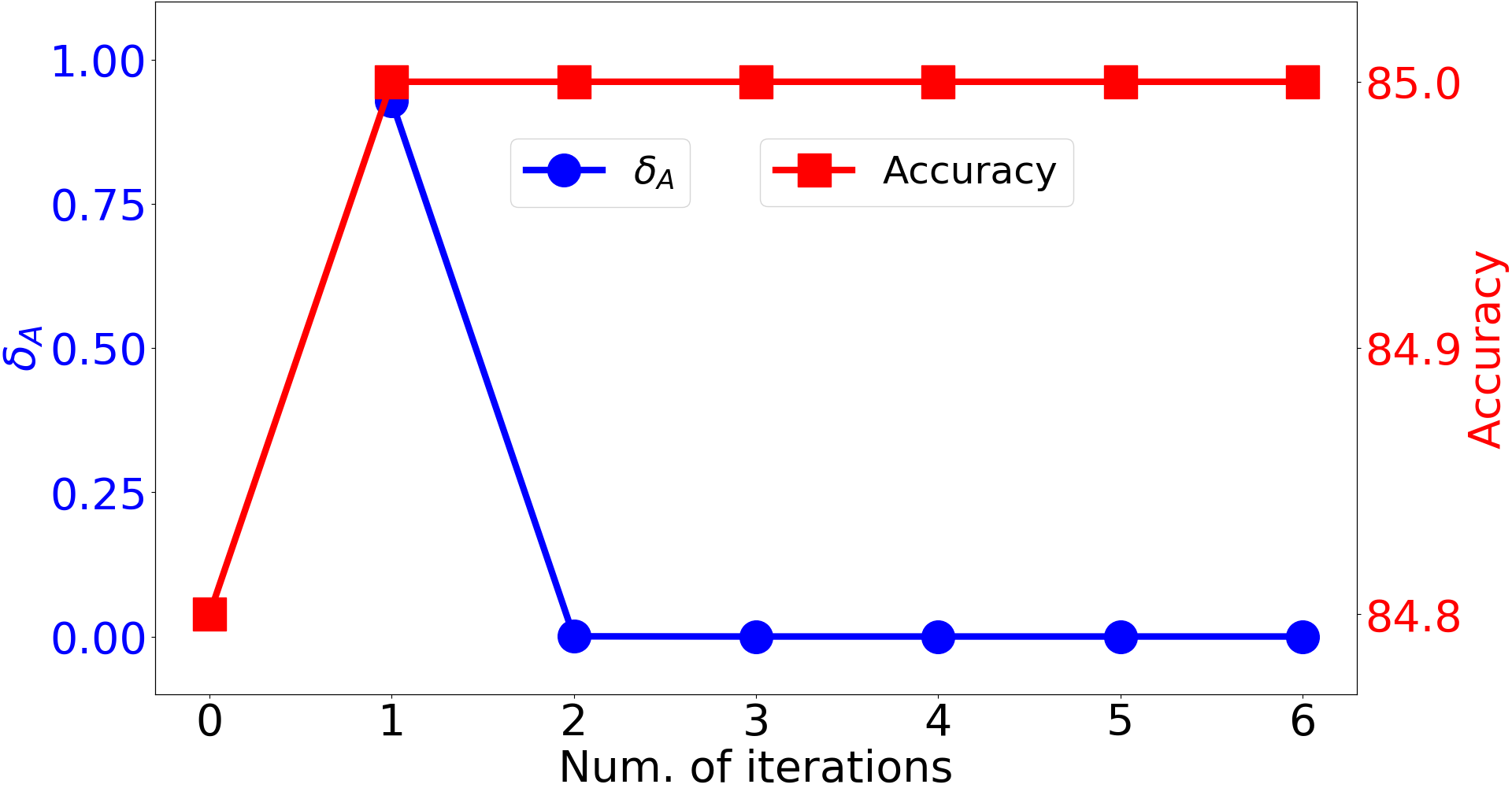}
	\label{fig:cora_conv}
	}
		  %\hspace{0.1in}
  \subfloat[t][
%   Performance comparison (i.e., test accuracy in \%) of two different stopping strategies on Cora.
  Stopping strategy comparison on Cora.
  ]{%
    \includegraphics[keepaspectratio=true,scale=0.14]{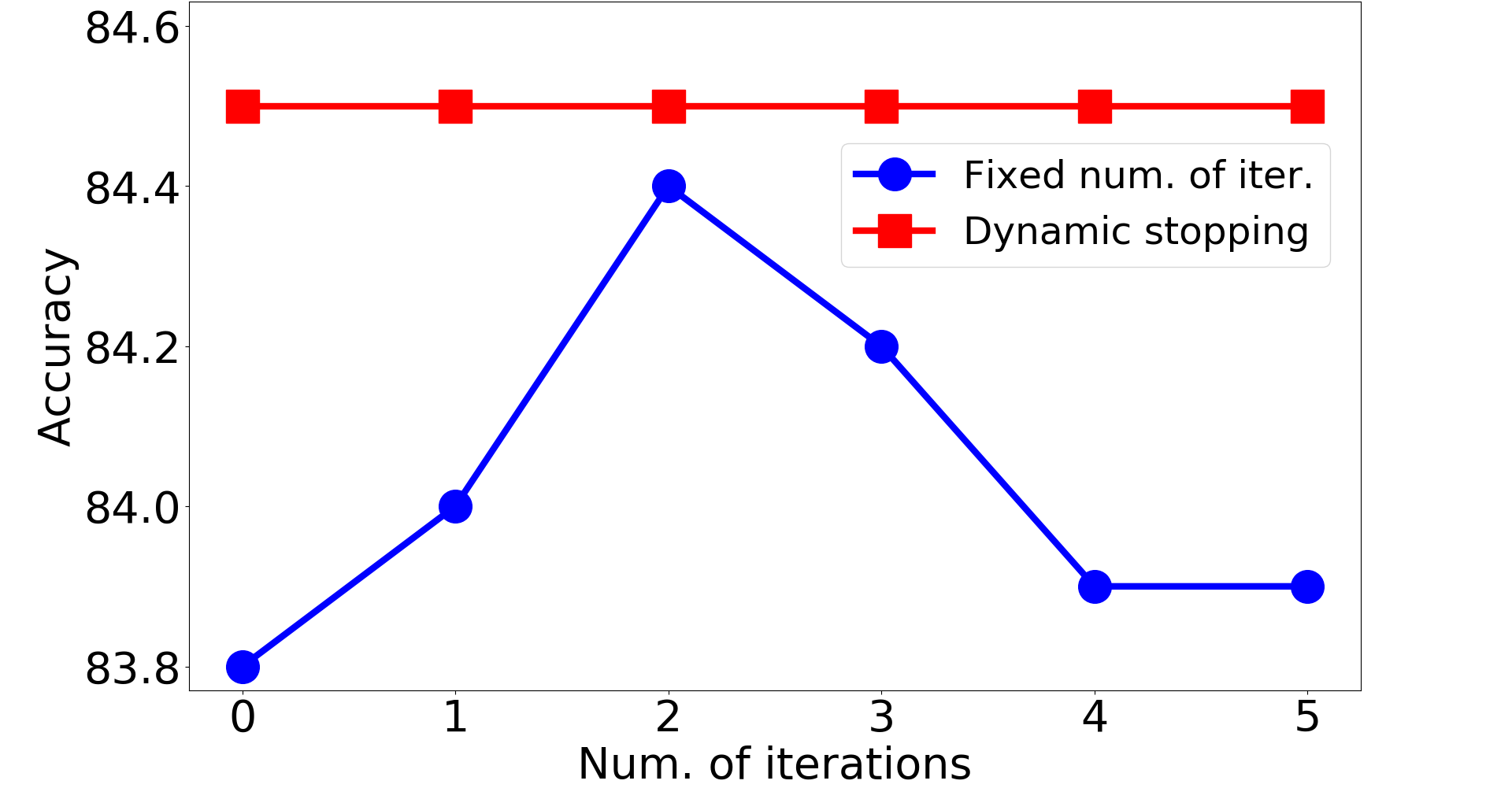}
	\label{fig:cora_stopping}
	}
 }
%   \vspace{-0.1in}
  \caption{Convergence and stopping strategy study on Cora (Single run results).}
  \label{fig:cora_convergence_stop_plot}
%   \vspace{-0.2in}
\end{figure}

In~\cref{fig:cora_conv} (and~\cref{sec:convergence_test}), we show the evolution of the learned adjacency matrix and accuracy through iterations in the iterative learning procedure in the testing phase.
We compute the difference between adjacency matrices at consecutive iterations as $\delta_A^{(t)} = ||\vec{A}^{(t)} - \vec{A}^{(t-1)}||_F^2 / ||\vec{A}^{(t)}||_F^2$ which typically ranges from 0 to 1.
As we can see, both the adjacency matrix and accuracy converge quickly.
This empirically verifies the analysis we made on the convergence property of \alg in~\cref{sec:convergence_analysis}.
Please note that this convergence property is not due to the oversmoothing effect of GNNs~\cite{xu2018representation,li2018deeper}, because we only employ a two-layered GCN as the underlying GNN module of \alg in our experiments.

% \noindent\textbf{Timing.}
We compare the training efficiency of \alg and \salg with other baselines.
% on various benchmarks. 
As shown in~\cref{table:training_time},
\alg is consistently faster than LDS, but in general, they are comparable.
Note that \alg has comparable model size compared to LDS. 
For instance, on the Cora data, the number of trainable parameters of \alg is 28,836, and for LDS, it is 23,040. 
And we see a large speedup of \salg compared to \alg.
Note that we were not able to run \alg on Pubmed because of memory limitation.
The theoretical complexity analysis is provided in~\cref{sec:model_complexity}.

We also empirically study the stopping strategy~(\cref{fig:cora_stopping} and~\cref{sec:stopping_strategy_analysis}), visualize the graph structures learned by \alg (\cref{sec:graph_viz}),
and conduct hyperparameter analysis (\cref{sec:hyperparam_analysis}).
Details on model settings are provided in~\cref{sec:model_settings}.

%  \begin{table}[tb]
\begin{wraptable}{r}{3in}
% \vspace{-1mm}
\caption{Mean and standard deviation of training time (5 runs) on various benchmarks (in seconds).}
\label{table:training_time}
\centering
% \addtolength{\tabcolsep}{-4.5pt}
\scalebox{0.95}{

\begin{tabular}{lllll}
\hline
  Data & \vline & Cora & Citeseer & Pubmed\\
  \hline
  GCN & \vline & \textbf{3 (1)} & \textbf{5 (1)} & \textbf{29 (4)}  \\ 
  GAT & \vline & 26 (5) & 28 (5)& \quad\textrm{---}  \\ 
  LDS  & \vline & 390 (82) &585 (181) & \quad\textrm{---}\\ 
  \hline
  \alg &\vline & 237 (21) & 563 (100) & \quad\textrm{---} \\ 
  w/o IL &\vline &49 (8) & 61 (15) &\quad\textrm{---}   \\
  \hline
    \salg &\vline & 83 (6)  & 261 (50) & 323 (53)\\ 
  w/o IL &\vline & 28 (4) &  69 (9) &  71 (17) \\
 \hline
\end{tabular}
}
% \vspace{-1mm}
% \end{table}
\end{wraptable}

\section{Related Work}
% key points:
% clearly define the problem
% thoroughly acknowledge related work
% precisely explain the relation with previous works

% Structure learning for Bayesian networks
% link prediction: check if existing work on joint learning

% \subsection{Graph Structure Learning}

The problem of graph structure learning has been widely studied in different fields from different perspectives.
In the field of graph signal processing,
researchers have explored various ways of learning graphs from data~\citep{dong2016learning,egilmez2017graph,wang2019graspel,kang2019robust,berger2020efficient,bai2020multi}, with certain structural constraints (e.g., sparsity) on the graphs.
This problem has also been studied in the literature of clustering analysis~\citep{bojchevski2017robust,huang2019auto} where they aimed to simultaneously perform the clustering task and learn similarity relationships among objects.
These works all focused on unsupervised learning setting without considering any supervised downstream tasks, and were incapable of handling inductive learning problems.
% no downstream tasks, transductive only
Other related works include structure inference in probabilistic graphical models~\citep{cussens2012bayesian,zheng2018dags,yu2019dag},
and graph generation~\citep{liao2019efficient,shi2020graphaf}, which have a different goal from ours.

% GNN: adersarial setting 
In the field of GNNs~\citep{kipf2016semi,gilmer2017neural,hamilton2017inductive,li2015gated,yun2019graph}, there is a line of research on developing robust GNNs~\citep{sun2018adversarial} that are invulnerable to adversarial graphs by leveraging 
% various techniques including
attention-based methods~\citep{chen2019label}, Bayesian methods~\citep{elinas2019variational,zhang2019bayesian}, graph diffusion-based methods~\citep{klicpera2019diffusion}, and various assumptions on graphs (e.g., low rank and sparsity)~\citep{entezari2020all,jin2020graph,zheng2020robust}.
These methods usually assume that the initial graph structure is available.
% GNN: automatic graph construction
% How to apply GNNs to applications where the underlying graph structures are unavailable becomes an emergent and challenging problem.
% However, manually constructing graphs from data heavily relies on domain knowledge and is not very scalable.
Recently, researchers have explored methods to automatically construct a graph 
of objects~\citep{norcliffe2018learning,choi2019graph,li2018adaptive,franceschi2019learning,liu2020automatic} or words~\citep{liu2018contextualized,chen2019graphflow,chen2019reinforcement} when applying GNNs to non-graph structured data.
% While most existing works focus on learning graph structures from homogeneous graphs, 
% ~\citep{yun2019graph} focused on graph learning from heterogeneous graphs.
% proposed to simultaneously learn meta-path based adjacency matrices and node embeddings from heterogeneous graphs.
However, these methods merely optimize the graphs toward the downstream tasks without the explicit control of the quality of the learned graphs.
More recently, \citep{franceschi2019learning} proposed the LDS model for jointly learning the graph and the parameters of GNNs by leveraging the bilevel optimization technique. 
% However, one big limitation of LDS is that it optimizes the discrete probability distribution on the edges of the graph,
However, by design, their method is unable to handle the inductive setting. 
Our work is also related to Transformer-like approaches~\citep{vaswani2017attention} that learn relationships among objects by leveraging multi-head attention mechanism.
However, these methods do not focus on the graph learning problem and were not designed to utilize the initial graph structure.

\section{Conclusion}

We proposed a novel \alg framework for jointly and iteratively learning the graph structure and embeddings optimized for the downstream task.
% The proposed method is able to iteratively search for an implicit graph structure that better helps the prediction task.
Experimental results demonstrate the effectiveness and efficiency of the proposed models. 
In the future, we plan to explore effective techniques for handling more challenging scenarios 
where both graph topology and node features are noisy.
% It will also be interesting to apply our framework to other GNN variants.

% \subsection{Margins in \LaTeX{}}

% Most of the margin problems come from figures positioned by hand using
% \verb+\special+ or other commands. We suggest using the command
% \verb+\includegraphics+ from the \verb+graphicx+ package. Always specify the
% figure width as a multiple of the line width as in the example below:
% \begin{verbatim}
%   \usepackage[pdftex]{graphicx} ...
%   \includegraphics[width=0.8\linewidth]{myfile.pdf}
% \end{verbatim}
% See Section 4.4 in the graphics bundle documentation
% (\url{http://mirrors.ctan.org/macros/latex/required/graphics/grfguide.pdf})

% A number of width problems arise when \LaTeX{} cannot properly hyphenate a
% line. Please give LaTeX hyphenation hints using the \verb+\-+ command when
% necessary.

% \clearpage
\section*{Broader Impact}

%% Applications
The fundamental goal of our research is to develop a method for jointly learning graph structures and embeddings that are optimized for (semi-)supervised downstream tasks.
Our technique can be widely applied to a large range of applications, including social network analysis, natural language processing (e.g., question answering and text generation), drug discovery and community detection.
Conceptually, any application with the purpose of jointly learning the graph structures and embeddings in order to perform well in downstream tasks.
Those potential applications range from computer vision, natural language processing, and network analysis.
For instance, our research might be used to help better capture the semantic relationships between word tokens (beyond a sequence of tokens) in natural language processing.

%% Implications
There are many benefits of using our method as a tool, such as applying graph neural networks to non-graph structured data without manual graph construction,
and learning node/graph embeddings that are more robust to noisy input graphs.
Those benefits that might be utilized by a large number of potential applications may have a board range of societal impacts:
\begin{itemize}
\item the use of our research could improve and speed up the process of learning meaningful graphs from noisy/incomplete graphs (e.g., social networks) or even non-graph structured data (e.g., text and images).

\item the use of our research could improve the robustness of graph neural networks to noisy/incomplete graph-structured data in terms of learning good node/graph embeddings for downstream task. 
\end{itemize}

%% Initiatives
We would encourage the research to explore similar approaches in more specific real-world applications.
We would also suggest the research to understand the adversarial robustness of the use of graph neural networks in safety/security-critical applications.

\begin{ack}
This research is sponsored by the Defense Advanced Research Projects Agency (DARPA) through Cooperative Agreement D20AC00004 awarded by the U.S. Department of the Interior (DOI), Interior Business Center. The content of the information does not necessarily reflect the position or the policy of the Government, and no official endorsement should be inferred. This work is also supported by IBM Research AI through the IBM AI Horizons Network. We thank the anonymous reviewers for their constructive suggestions.
\end{ack}

\bibliography{neurips_2020}

\begin{thebibliography}{10}

\bibitem{bai2020multi}
X.~Bai, L.~Zhu, C.~Liang, J.~Li, X.~Nie, and X.~Chang.
\newblock Multi-view feature selection via nonnegative structured graph
  learning.
\newblock {\em Neurocomputing}, 2020.

\bibitem{belkin2002laplacian}
M.~Belkin and P.~Niyogi.
\newblock Laplacian eigenmaps and spectral techniques for embedding and
  clustering.
\newblock In {\em Advances in neural information processing systems}, pages
  585--591, 2002.

\bibitem{berger2020efficient}
P.~Berger, G.~Hannak, and G.~Matz.
\newblock Efficient graph learning from noisy and incomplete data.
\newblock {\em IEEE Transactions on Signal and Information Processing over
  Networks}, 6:105--119, 2020.

\bibitem{bojchevski2017robust}
A.~Bojchevski, Y.~Matkovic, and S.~G{\"u}nnemann.
\newblock Robust spectral clustering for noisy data: Modeling sparse
  corruptions improves latent embeddings.
\newblock In {\em Proceedings of the 23rd ACM SIGKDD International Conference
  on Knowledge Discovery and Data Mining}, pages 737--746, 2017.

\bibitem{chen2019label}
H.~Chen, L.~Wang, S.~Wang, D.~Luo, W.~Huang, and Z.~Li.
\newblock Label aware graph convolutional network--not all edges deserve your
  attention.
\newblock {\em arXiv preprint arXiv:1907.04707}, 2019.

\bibitem{chen2019graphflow}
Y.~Chen, L.~Wu, and M.~J. Zaki.
\newblock Graphflow: Exploiting conversation flow with graph neural networks
  for conversational machine comprehension.
\newblock {\em arXiv preprint arXiv:1908.00059}, 2019.

\bibitem{chen2019reinforcement}
Y.~Chen, L.~Wu, and M.~J. Zaki.
\newblock Reinforcement learning based graph-to-sequence model for natural
  question generation.
\newblock {\em arXiv preprint arXiv:1908.04942}, 2019.

\bibitem{choi2019graph}
E.~Choi, Z.~Xu, Y.~Li, M.~W. Dusenberry, G.~Flores, Y.~Xue, and A.~M. Dai.
\newblock Graph convolutional transformer: Learning the graphical structure of
  electronic health records.
\newblock {\em arXiv preprint arXiv:1906.04716}, 2019.

\bibitem{cussens2012bayesian}
J.~Cussens.
\newblock Bayesian network learning with cutting planes.
\newblock {\em arXiv preprint arXiv:1202.3713}, 2012.

\bibitem{dong2016learning}
X.~Dong, D.~Thanou, P.~Frossard, and P.~Vandergheynst.
\newblock Learning laplacian matrix in smooth graph signal representations.
\newblock {\em IEEE Transactions on Signal Processing}, 64(23):6160--6173,
  2016.

\bibitem{Dua:2019}
D.~Dua and C.~Graff.
\newblock {UCI} machine learning repository, 2017.

\bibitem{egilmez2017graph}
H.~E. Egilmez, E.~Pavez, and A.~Ortega.
\newblock Graph learning from data under laplacian and structural constraints.
\newblock {\em IEEE Journal of Selected Topics in Signal Processing},
  11(6):825--841, 2017.

\bibitem{elinas2019variational}
P.~Elinas, E.~V. Bonilla, and L.~Tiao.
\newblock Variational inference for graph convolutional networks in the absence
  of graph data and adversarial settings.
\newblock {\em arXiv}, pages arXiv--1906, 2019.

\bibitem{entezari2020all}
N.~Entezari, S.~A. Al-Sayouri, A.~Darvishzadeh, and E.~E. Papalexakis.
\newblock All you need is low (rank) defending against adversarial attacks on
  graphs.
\newblock In {\em Proceedings of the 13th International Conference on Web
  Search and Data Mining}, pages 169--177, 2020.

\bibitem{franceschi2019learning}
L.~Franceschi, M.~Niepert, M.~Pontil, and X.~He.
\newblock Learning discrete structures for graph neural networks.
\newblock {\em arXiv preprint arXiv:1903.11960}, 2019.

\bibitem{gilmer2017neural}
J.~Gilmer, S.~S. Schoenholz, P.~F. Riley, O.~Vinyals, and G.~E. Dahl.
\newblock Neural message passing for quantum chemistry.
\newblock In {\em Proceedings of the 34th International Conference on Machine
  Learning-Volume 70}, pages 1263--1272. JMLR. org, 2017.

\bibitem{GraphSage:hamilton2017inductive}
W.~Hamilton, Z.~Ying, and J.~Leskovec.
\newblock Inductive representation learning on large graphs.
\newblock In {\em Advances in Neural Information Processing Systems}, 2017.

\bibitem{hamilton2017inductive}
W.~Hamilton, Z.~Ying, and J.~Leskovec.
\newblock Inductive representation learning on large graphs.
\newblock In {\em Advances in Neural Information Processing Systems}, pages
  1024--1034, 2017.

\bibitem{hochreiter1997long}
S.~Hochreiter and J.~Schmidhuber.
\newblock Long short-term memory.
\newblock {\em Neural computation}, 9(8):1735--1780, 1997.

\bibitem{hu2019semi}
F.~Hu, Y.~Zhu, S.~Wu, L.~Wang, and T.~Tan.
\newblock Semi-supervised node classification via hierarchical graph
  convolutional networks.
\newblock {\em arXiv preprint arXiv:1902.06667}, 2019.

\bibitem{hu2020open}
W.~Hu, M.~Fey, M.~Zitnik, Y.~Dong, H.~Ren, B.~Liu, M.~Catasta, and J.~Leskovec.
\newblock Open graph benchmark: Datasets for machine learning on graphs.
\newblock {\em arXiv preprint arXiv:2005.00687}, 2020.

\bibitem{huang2019auto}
S.~Huang, Z.~Kang, I.~W. Tsang, and Z.~Xu.
\newblock Auto-weighted multi-view clustering via kernelized graph learning.
\newblock {\em Pattern Recognition}, 88:174--184, 2019.

\bibitem{jiang2019semi}
B.~Jiang, Z.~Zhang, D.~Lin, J.~Tang, and B.~Luo.
\newblock Semi-supervised learning with graph learning-convolutional networks.
\newblock In {\em Proceedings of the IEEE Conference on Computer Vision and
  Pattern Recognition}, pages 11313--11320, 2019.

\bibitem{jin2020graph}
W.~Jin, Y.~Ma, X.~Liu, X.~Tang, S.~Wang, and J.~Tang.
\newblock Graph structure learning for robust graph neural networks.
\newblock {\em arXiv preprint arXiv:2005.10203}, 2020.

\bibitem{kalofolias2016learn}
V.~Kalofolias.
\newblock How to learn a graph from smooth signals.
\newblock In {\em Artificial Intelligence and Statistics}, pages 920--929,
  2016.

\bibitem{kalofolias2017large}
V.~Kalofolias and N.~Perraudin.
\newblock Large scale graph learning from smooth signals.
\newblock {\em arXiv preprint arXiv:1710.05654}, 2017.

\bibitem{kang2019robust}
Z.~Kang, H.~Pan, S.~C. Hoi, and Z.~Xu.
\newblock Robust graph learning from noisy data.
\newblock {\em IEEE transactions on cybernetics}, 2019.

\bibitem{kingma2014adam}
D.~P. Kingma and J.~Ba.
\newblock Adam: A method for stochastic optimization.
\newblock {\em arXiv preprint arXiv:1412.6980}, 2014.

\bibitem{kipf2016semi}
T.~N. Kipf and M.~Welling.
\newblock Semi-supervised classification with graph convolutional networks.
\newblock {\em arXiv preprint arXiv:1609.02907}, 2016.

\bibitem{klicpera2018predict}
J.~Klicpera, A.~Bojchevski, and S.~G{\"u}nnemann.
\newblock Predict then propagate: Graph neural networks meet personalized
  pagerank.
\newblock {\em arXiv preprint arXiv:1810.05997}, 2018.

\bibitem{klicpera2019diffusion}
J.~Klicpera, S.~Wei{\ss}enberger, and S.~G{\"u}nnemann.
\newblock Diffusion improves graph learning.
\newblock In {\em Advances in Neural Information Processing Systems}, pages
  13333--13345, 2019.

\bibitem{lang1995newsweeder}
K.~Lang.
\newblock Newsweeder: Learning to filter netnews.
\newblock In {\em Machine Learning Proceedings 1995}, pages 331--339. Elsevier,
  1995.

\bibitem{li2018deeper}
Q.~Li, Z.~Han, and X.-M. Wu.
\newblock Deeper insights into graph convolutional networks for semi-supervised
  learning.
\newblock In {\em Thirty-Second AAAI Conference on Artificial Intelligence},
  2018.

\bibitem{li2018adaptive}
R.~Li, S.~Wang, F.~Zhu, and J.~Huang.
\newblock Adaptive graph convolutional neural networks.
\newblock In {\em Thirty-Second AAAI Conference on Artificial Intelligence},
  2018.

\bibitem{li2015gated}
Y.~Li, D.~Tarlow, M.~Brockschmidt, and R.~Zemel.
\newblock Gated graph sequence neural networks.
\newblock {\em arXiv preprint arXiv:1511.05493}, 2015.

\bibitem{li2016gated}
Y.~Li, D.~Tarlow, M.~Brockschmidt, and R.~Zemel.
\newblock Gated graph sequence neural networks.
\newblock {\em International Conference on Learning Representations}, 2016.

\bibitem{li2018learning}
Y.~Li, O.~Vinyals, C.~Dyer, R.~Pascanu, and P.~Battaglia.
\newblock Learning deep generative models of graphs.
\newblock {\em arXiv preprint arXiv:1803.03324}, 2018.

\bibitem{liao2019efficient}
R.~Liao, Y.~Li, Y.~Song, S.~Wang, W.~Hamilton, D.~K. Duvenaud, R.~Urtasun, and
  R.~Zemel.
\newblock Efficient graph generation with graph recurrent attention networks.
\newblock In {\em Advances in Neural Information Processing Systems}, pages
  4257--4267, 2019.

\bibitem{liu2018contextualized}
P.~Liu, S.~Chang, X.~Huang, J.~Tang, and J.~C.~K. Cheung.
\newblock Contextualized non-local neural networks for sequence learning.
\newblock {\em arXiv preprint arXiv:1811.08600}, 2018.

\bibitem{liu2020automatic}
S.~Liu, Y.~Chen, X.~Xie, J.~K. Siow, and Y.~Liu.
\newblock Automatic code summarization via multi-dimensional semantic fusing in
  gnn.
\newblock {\em arXiv preprint arXiv:2006.05405}, 2020.

\bibitem{liu2010large}
W.~Liu, J.~He, and S.-F. Chang.
\newblock Large graph construction for scalable semi-supervised learning.
\newblock In {\em ICML}, 2010.

\bibitem{lovasz1993random}
L.~Lov{\'a}sz.
\newblock Random walks on graphs: A survey.
\newblock Department of Computer Science, Yale University, 1994.

\bibitem{ma2019graph}
Y.~Ma, S.~Wang, C.~C. Aggarwal, and J.~Tang.
\newblock Graph convolutional networks with eigenpooling.
\newblock {\em arXiv preprint arXiv:1904.13107}, 2019.

\bibitem{nguyen2010cosine}
H.~V. Nguyen and L.~Bai.
\newblock Cosine similarity metric learning for face verification.
\newblock In {\em Asian conference on computer vision}, pages 709--720.
  Springer, 2010.

\bibitem{norcliffe2018learning}
W.~Norcliffe-Brown, S.~Vafeias, and S.~Parisot.
\newblock Learning conditioned graph structures for interpretable visual
  question answering.
\newblock In {\em Advances in Neural Information Processing Systems}, pages
  8344--8353, 2018.

\bibitem{pang2004sentimental}
B.~Pang and L.~Lee.
\newblock A sentimental education: Sentiment analysis using subjectivity
  summarization based on minimum cuts.
\newblock In {\em Proceedings of the 42nd annual meeting on Association for
  Computational Linguistics}, page 271. Association for Computational
  Linguistics, 2004.

\bibitem{samanta2018designing}
B.~Samanta, A.~De, N.~Ganguly, and M.~Gomez-Rodriguez.
\newblock Designing random graph models using variational autoencoders with
  applications to chemical design.
\newblock {\em arXiv preprint arXiv:1802.05283}, 2018.

\bibitem{sen2008collective}
P.~Sen, G.~Namata, M.~Bilgic, L.~Getoor, B.~Galligher, and T.~Eliassi-Rad.
\newblock Collective classification in network data.
\newblock {\em AI magazine}, 29(3):93--93, 2008.

\bibitem{shi2020graphaf}
C.~Shi, M.~Xu, Z.~Zhu, W.~Zhang, M.~Zhang, and J.~Tang.
\newblock Graphaf: a flow-based autoregressive model for molecular graph
  generation.
\newblock {\em arXiv preprint arXiv:2001.09382}, 2020.

\bibitem{sun2018adversarial}
L.~Sun, Y.~Dou, C.~Yang, J.~Wang, P.~S. Yu, and B.~Li.
\newblock Adversarial attack and defense on graph data: A survey.
\newblock {\em arXiv preprint arXiv:1812.10528}, 2018.

\bibitem{vaswani2017attention}
A.~Vaswani, N.~Shazeer, N.~Parmar, J.~Uszkoreit, L.~Jones, A.~N. Gomez,
  {\L}.~Kaiser, and I.~Polosukhin.
\newblock Attention is all you need.
\newblock In {\em Advances in neural information processing systems}, pages
  5998--6008, 2017.

\bibitem{velivckovic2017graph}
P.~Veli{\v{c}}kovi{\'c}, G.~Cucurull, A.~Casanova, A.~Romero, P.~Li{\`o}, and
  Y.~Bengio.
\newblock Graph attention networks.
\newblock {\em arXiv preprint arXiv:1710.10903}, 2017.

\bibitem{wang2019graspel}
Y.~Wang, Z.~Zhao, and Z.~Feng.
\newblock Graspel: Graph spectral learning at scale.
\newblock {\em arXiv preprint arXiv:1911.10373}, 2019.

\bibitem{wojke2018deep}
N.~Wojke and A.~Bewley.
\newblock Deep cosine metric learning for person re-identification.
\newblock In {\em 2018 IEEE winter conference on applications of computer
  vision (WACV)}, pages 748--756. IEEE, 2018.

\bibitem{wu2019scalable}
L.~Wu, I.~E.-H. Yen, Z.~Zhang, K.~Xu, L.~Zhao, X.~Peng, Y.~Xia, and
  C.~Aggarwal.
\newblock Scalable global alignment graph kernel using random features: From
  node embedding to graph embedding.
\newblock In {\em Proceedings of the 25th ACM SIGKDD International Conference
  on Knowledge Discovery \& Data Mining}, pages 1418--1428, 2019.

\bibitem{xu2018representation}
K.~Xu, C.~Li, Y.~Tian, T.~Sonobe, K.-i. Kawarabayashi, and S.~Jegelka.
\newblock Representation learning on graphs with jumping knowledge networks.
\newblock {\em arXiv preprint arXiv:1806.03536}, 2018.

\bibitem{xu2018graph2seq}
K.~Xu, L.~Wu, Z.~Wang, and V.~Sheinin.
\newblock Graph2seq: Graph to sequence learning with attention-based neural
  networks.
\newblock {\em arXiv preprint arXiv:1804.00823}, 2018.

\bibitem{xu2018exploiting}
K.~Xu, L.~Wu, Z.~Wang, M.~Yu, L.~Chen, and V.~Sheinin.
\newblock Exploiting rich syntactic information for semantic parsing with
  graph-to-sequence model.
\newblock {\em arXiv preprint arXiv:1808.07624}, 2018.

\bibitem{yeung2007kernel}
D.-Y. Yeung and H.~Chang.
\newblock A kernel approach for semisupervised metric learning.
\newblock {\em IEEE Transactions on Neural Networks}, 18(1):141--149, 2007.

\bibitem{ying2018hierarchical}
Z.~Ying, J.~You, C.~Morris, X.~Ren, W.~Hamilton, and J.~Leskovec.
\newblock Hierarchical graph representation learning with differentiable
  pooling.
\newblock In {\em Advances in Neural Information Processing Systems}, pages
  4800--4810, 2018.

\bibitem{you2018graphrnn}
J.~You, R.~Ying, X.~Ren, W.~L. Hamilton, and J.~Leskovec.
\newblock Graphrnn: Generating realistic graphs with deep auto-regressive
  models.
\newblock {\em arXiv preprint arXiv:1802.08773}, 2018.

\bibitem{yu2019dag}
Y.~Yu, J.~Chen, T.~Gao, and M.~Yu.
\newblock Dag-gnn: Dag structure learning with graph neural networks.
\newblock {\em arXiv preprint arXiv:1904.10098}, 2019.

\bibitem{yun2019graph}
S.~Yun, M.~Jeong, R.~Kim, J.~Kang, and H.~J. Kim.
\newblock Graph transformer networks.
\newblock In {\em Advances in Neural Information Processing Systems}, pages
  11960--11970, 2019.

\bibitem{zhang2019bayesian}
Y.~Zhang, S.~Pal, M.~Coates, and D.~Ustebay.
\newblock Bayesian graph convolutional neural networks for semi-supervised
  classification.
\newblock In {\em Proceedings of the AAAI Conference on Artificial
  Intelligence}, volume~33, pages 5829--5836, 2019.

\bibitem{zheng2020robust}
C.~Zheng, B.~Zong, W.~Cheng, D.~Song, J.~Ni, W.~Yu, H.~Chen, and W.~Wang.
\newblock Robust graph representation learning via neural sparsification.
\newblock In {\em ICML}, 2020.

\bibitem{zheng2018dags}
X.~Zheng, B.~Aragam, P.~K. Ravikumar, and E.~P. Xing.
\newblock Dags with no tears: Continuous optimization for structure learning.
\newblock In {\em Advances in Neural Information Processing Systems}, pages
  9472--9483, 2018.

\end{thebibliography}
\bibliographystyle{abbrv}

\clearpage
\appendix

% \section{Full Algorithm of the \alg and \salg Models}\label{sec:full_algorithm}

\section{Theoretical Model Analysis}

\subsection{Theoretical Proof of Recovering Node and Anchor Graphs from Affinity Matrix $\vec{R}$}\label{sec:proof_on_two_step_random_walks}

It is worth noting that a node-anchor affinity matrix $\vec{R}$ serves as a weighted adjacency matrix of a bipartite graph $\mathcal{B}$.
We hence establish stationary Markov random walks~\citep{lovasz1993random} by defining the one-step transition probabilities as follows,
\begin{equation}\label{eq:node_anchor_one_step_random_walk}
\begin{aligned}
p^{(1)}(u_k|v_i) &= \frac{R_{ik}}{\sum_{k'=1}^{s}{R_{ik'}}},\quad
p^{(1)}(v_i|u_k) &= \frac{R_{ik}}{\sum_{i'=1}^{n}{R_{i'k}}},\quad
\forall v_i \in \mathcal{V},\quad \forall u_k \in \mathcal{U}
\end{aligned}
\end{equation}

We can further compute the two-step transition probabilities between nodes as follows,
\begin{equation}
\begin{aligned}
p^{(2)}(v_j|v_i) &= \sum_{k=1}^{s}{p^{(1)}(v_j|u_k) p^{(1)}(u_k|v_i)}=
\sum_{k=1}^{s}{\frac{R_{jk}}{\sum_{j'=1}^{n}{R_{j'k}}} \frac{R_{ik}}{\sum_{k'=1}^{s}{R_{ik'}}}}=\sum_{k=1}^{s}{\frac{R_{jk}}{\Lambda_{kk}} \frac{R_{ik}}{\Delta_{ii}}}
\end{aligned}
\end{equation}
where $\Lambda_{kk}=\sum_{j'=1}^{n}{R_{j'k}}$ and $\Delta_{ii}=\sum_{k'=1}^{s}{R_{ik'}}$.
Therefore, we can recover a row-normalized adjacency matrix $\vec{A} \in \mathbb{R}^{n \times n}$ for the node graph as 
$A_{ij}=p^{(2)}(v_j|v_i)$, 
which can be further written in a compact form $\vec{A}= \vec{\Delta}^{-1} \vec{R} \vec{\Lambda}^{-1} \vec{R}^\top$.

Similarly, we can compute the two-step transition probabilities between anchors as follows,
\begin{equation}
\begin{aligned}
p^{(2)}(u_r|u_k) &= \sum_{i=1}^{n}{p^{(1)}(u_r|v_i) p^{(1)}(v_i|u_k)}=
\sum_{i=1}^{n}{\frac{R_{ir}}{\sum_{r'=1}^{s}{R_{ir'}}} \frac{R_{ik}}{\sum_{i'=1}^{n}{R_{i'k}}}}=\sum_{i=1}^{n}{\frac{R_{ir}}{\Delta_{ii}} \frac{R_{ik}}{\Lambda_{kk}}}
\end{aligned}
\end{equation}
And a row-normalized adjacency matrix $\vec{B} \in \mathbb{R}^{s \times s}$ for the anchor graph $\mathcal{Q}$ can be formulated as 
$B_{kr}=p^{(2)}(u_r|u_k)$.
And we can obtain $\vec{B}= \vec{\Lambda}^{-1} \vec{R}^\top \vec{\Delta}^{-1} \vec{R}$.

\subsection{Theoretical Convergence Analysis}\label{sec:convergence_analysis}

% \subsubsection{Convergence Analysis}\label{sec:convergence_analysis}

\begin{figure}[!htb]
% \vspace{-1mm}
  \centering
    \includegraphics[keepaspectratio=true,scale=0.25]{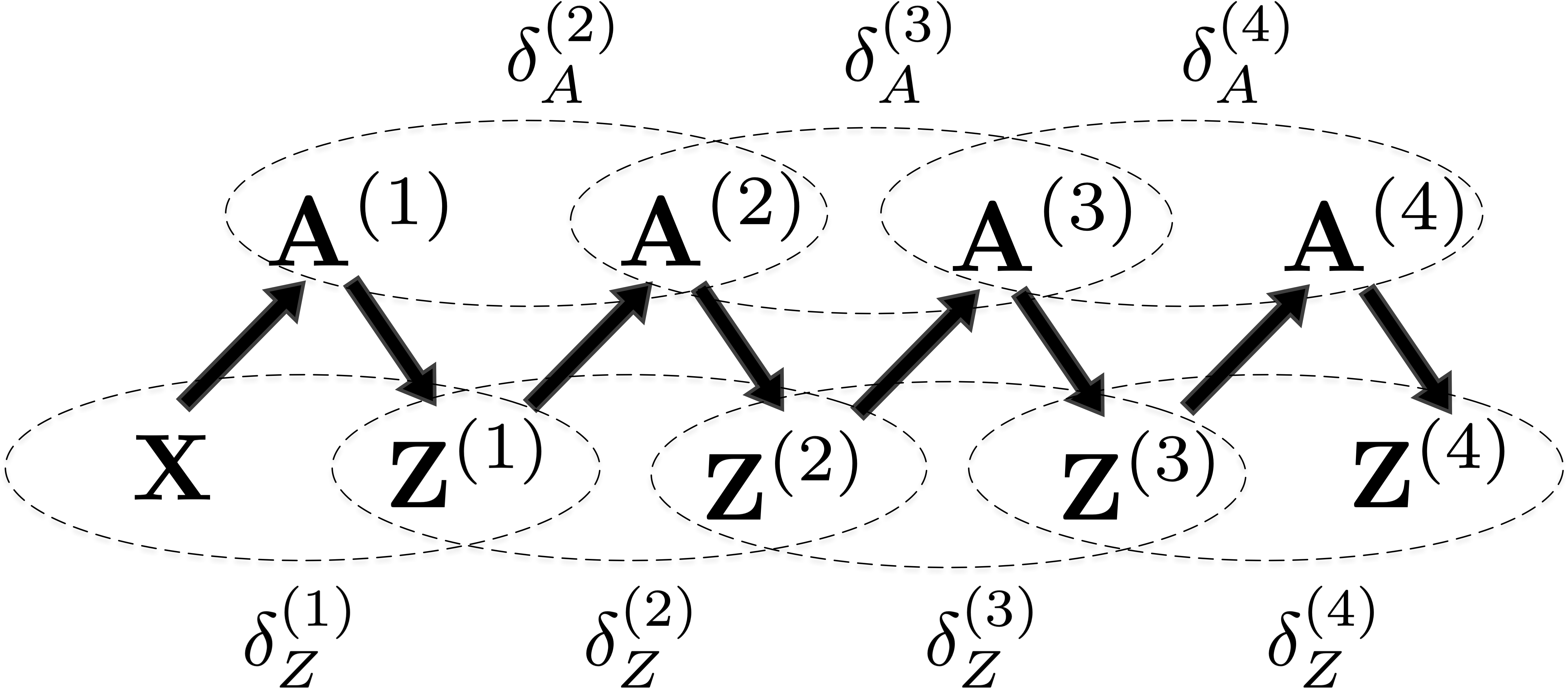}
  \caption{Information flow of iterative learning procedure.}
  \label{fig:info_flow}
\end{figure}

While it is challenging to theoretically prove the convergence of the proposed iterative learning procedure due to the arbitrary complexity of the  model, here we want to conceptually understand why it works in practice.
\cref{fig:info_flow} shows the information flow of the learned adjacency matrix $\vec{A}$ and the updated node embedding matrix $\vec{Z}$ during the iterative procedure. 
For the sake of simplicity, we omit some other variables such as $\widetilde{\vec{A}}$.
As we can see, at $t$-th iteration, $\vec{A}^{(t)}$ is computed based on $\vec{Z}^{(t-1)}$ (Line~\ref{alg_line:refine_adj}),
and $\vec{Z}^{(t)}$ is computed based on $\widetilde{\vec{A}}^{(t)}$ (Line~\ref{alg_line:refine_node_vec}) which is computed based on $\vec{A}^{(t)}$ (\cref{eq:combine_adj_norm_t}).
We further denote the difference between the adjacency matrices at the $t$-th iteration and the previous iteration by $\delta_A^{(t)}$.
Similarly, we denote the difference between the node embedding matrices at the $t$-th iteration and the previous iteration by $\delta_Z^{(t)}$.

If we assume that $\delta_Z^{(2)} < \delta_Z^{(1)}$, then we can expect that
$\delta_A^{(3)} < \delta_A^{(2)}$ because conceptually a more similar node embedding matrix (i.e., smaller $\delta_Z$) is supposed to produce a more similar adjacency matrix (i.e., smaller $\delta_A$) given the fact that model parameters keep the same through iterations.
Similarly, given that $\delta_A^{(3)} < \delta_A^{(2)}$, we can expect that  $\delta_Z^{(3)} < \delta_Z^{(2)}$.
Following this chain of reasoning, we can easily extend it to later iterations.
In order to see why the assumption $\delta_Z^{(2)} < \delta_Z^{(1)}$ makes sense in practice, we need to recall the fact that $\delta_Z^{(1)}$ measures the difference between $\vec{Z}^{(1)}$ and $\vec{X}$, which is usually larger than the difference between $\vec{Z}^{(2)}$ and $\vec{Z}^{(1)}$, namely $\delta_Z^{(2)}$.
For example, the raw node feature matrix $\vec{X}$ can be quite sparse in practice (e.g., in Cora and Citeseer), 
whereas $\vec{Z}^{(1)}$ is typically a dense matrix.
% We will empirically examine the convergence property of the iterative learning procedure in the experimental section.

\subsection{Model Complexity Analysis}\label{sec:model_complexity}

As for \alg, the cost of learning an adjacency matrix is $\mathcal{O}(n^2h)$ for $n$ nodes and data in $\mathbb{R}^h$, while computing node embeddings costs $\mathcal{O}(n^2h + ndh)$, 
computing task output costs $\mathcal{O}(n^2d)$, 
and computing the total loss costs $\mathcal{O}(n^2d)$ where $d$ is the hidden size.
We set the maximal number of iterations to $T$, hence
the overall complexity is $\mathcal{O}(Tn(nh+nd+hd))$.
If we assume that $d \approx h$ and $n \gg d$, the overall time complexity is $\mathcal{O}(Tdn^2)$.

As for \salg, the cost of learning a node-anchor affinity matrix is $\mathcal{O}(nsh)$,
while computing node embeddings costs $\mathcal{O}(nsh + ndh +|\mathcal{E}|h)$, 
computing task output costs $\mathcal{O}(nsd+|\mathcal{E}|d)$,
and computing the total loss costs $\mathcal{O}(ns^2 + s^2d)$ where $|\mathcal{E}|$ is the number of edges in the initial or kNN graph $\mathcal{G}$.
With the assumption that the initial or kNN graph is usually very sparse in practice, especially for large graphs, we hence set $|\mathcal{E}|=kn$ where $k$ is a constant denoting the average degree of the initial or kNN graph.
Therefore, we get the overall time complexity $\mathcal{O}(Tn(ds+d^2+s^2))$.
If we assume that $n \gg s$ which usually holds true for large graphs, the overall time complexity is linear with respect to the numbers of graph nodes $n$.

As for space complexity, compared to \alg, \salg reduces it from $\mathcal{O}(n^2)$ to $\mathcal{O}(ns)$ since it only needs to store the $n \times s$ affinity matrix.

\section{Empirical Model Analysis}\label{sec:more_model_analysis}

\subsection{Convergence Test}\label{sec:convergence_test}
Here, we show the evolution of the learned adjacency matrix and accuracy through iterations in the iterative learning procedure in the testing phase.
As we can see, both the adjacency matrix and accuracy converge quickly.

\begin{figure}[!htb]
  \centering
    \includegraphics[keepaspectratio=true,scale=0.16]{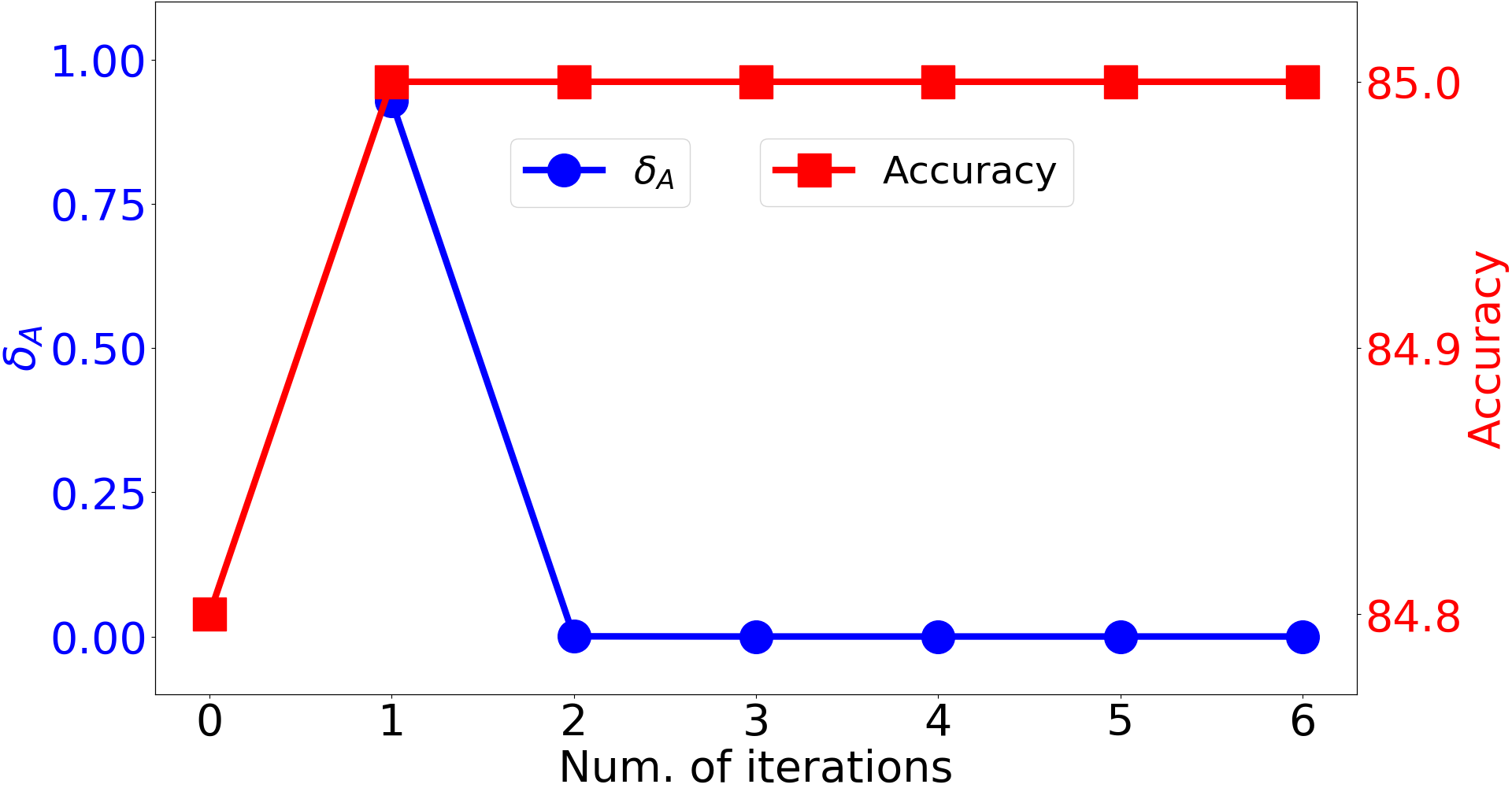}
    \caption{Convergence study on Cora (single run results).}
\end{figure}

\begin{figure}[!htb]
  \centering
    \includegraphics[keepaspectratio=true,scale=0.16]{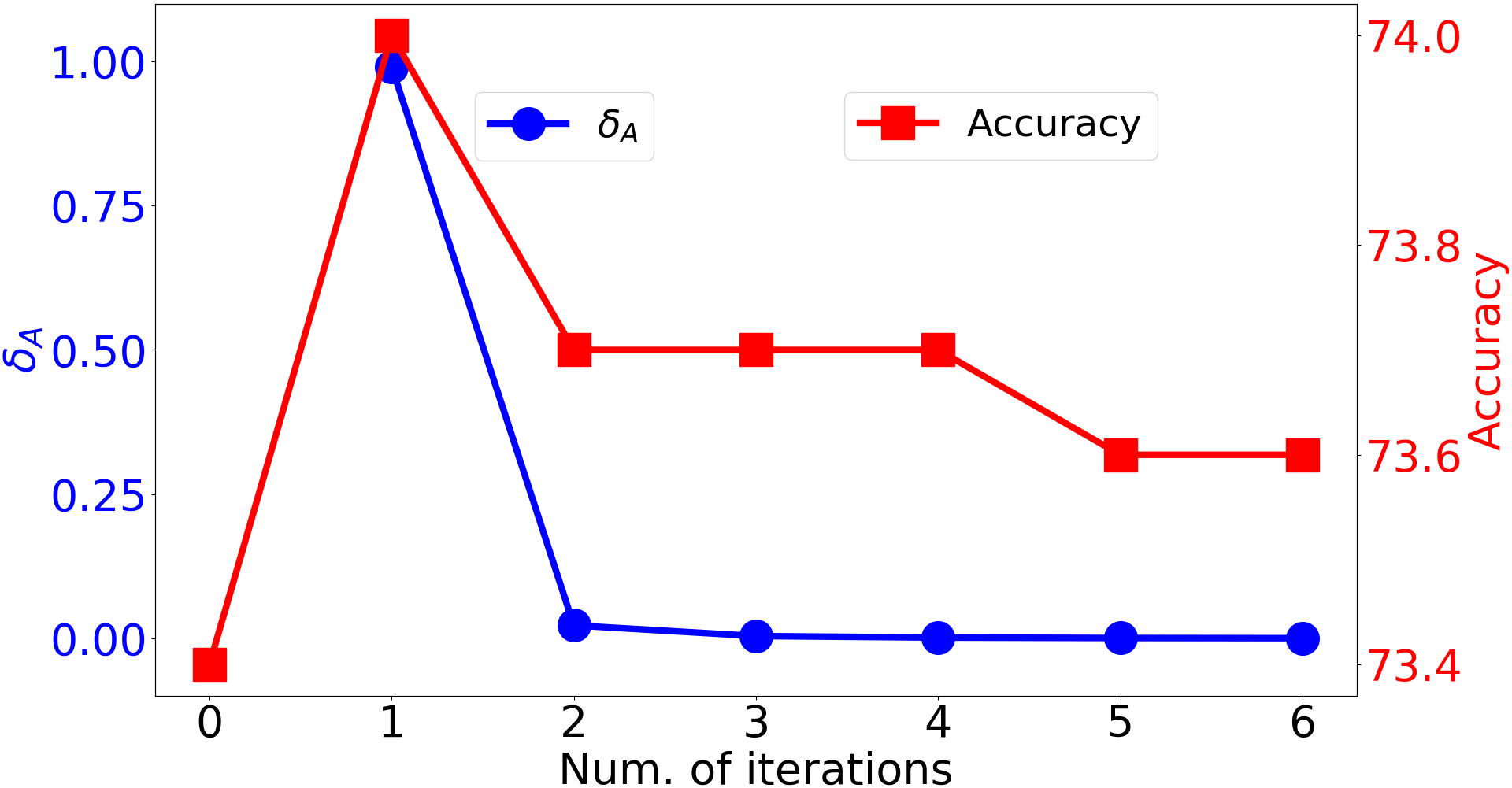}
    \caption{Convergence study on Citeseer (single run results).}
\end{figure}

% Stopping
\subsection{Stopping Strategy Analysis}\label{sec:stopping_strategy_analysis}
Here, we empirically compare the effectiveness of two stopping strategies: i) using a fixed number of iterations (blue line), and ii) using a stopping criterion to dynamically determine the convergence (red line).
As we can see, dynamically adjusting the number of iterations using the stopping criterion works better in practice.
Compared to using a fixed number of iterations globally, the advantage of applying this dynamical stopping strategy becomes more clear when we are doing mini-batch training since we can adjust when to stop dynamically for each example graph in the mini-batch.

\begin{figure}[!htb]
  \centering
        \includegraphics[keepaspectratio=true,scale=0.16]{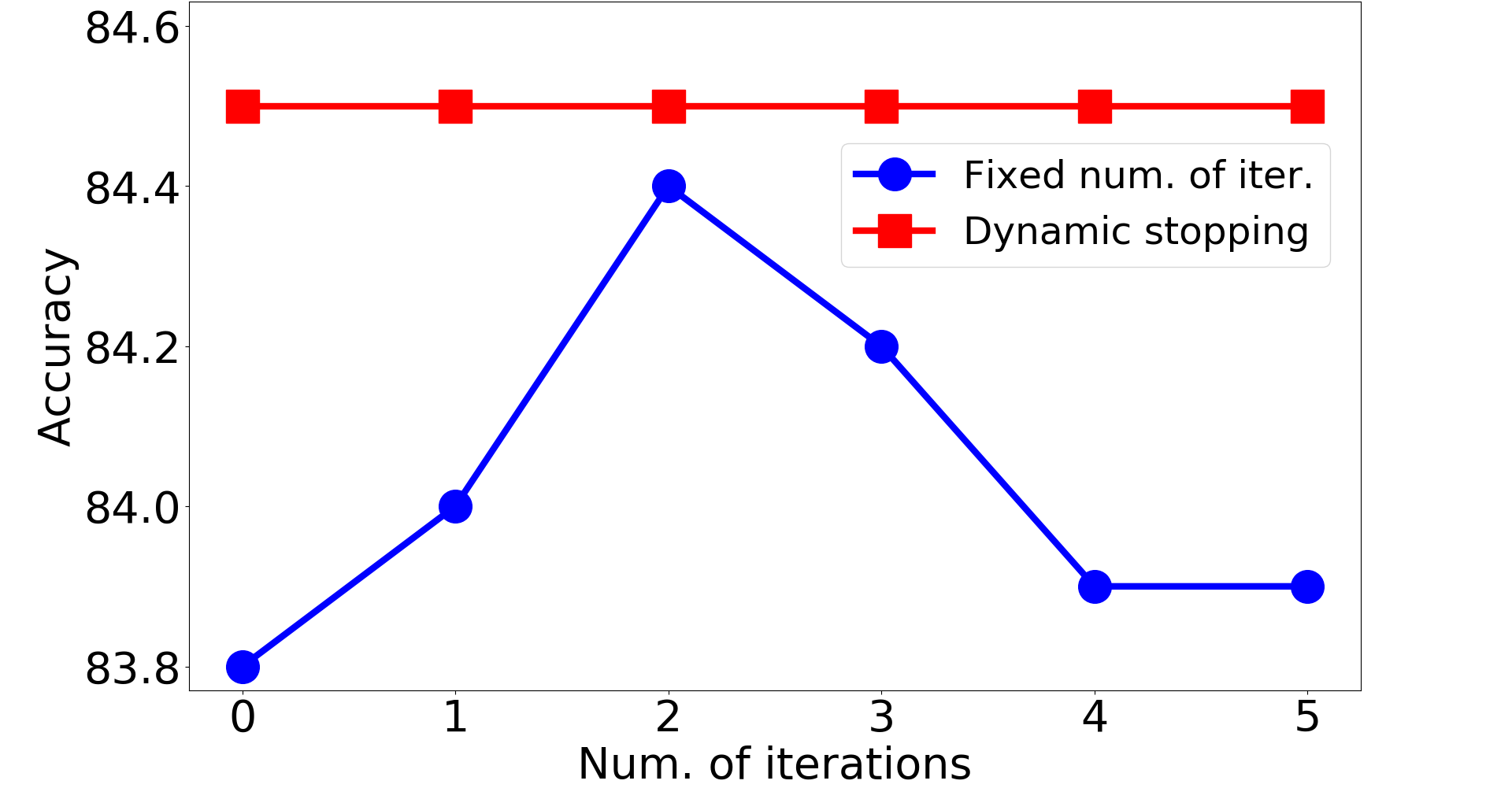}
    \caption{Performance comparison (i.e., test accuracy in \%) of two different stopping strategies on Cora.}
%   \label{fig:cora_stopping}
\end{figure}

\begin{figure}[!htb]
  \centering
        \includegraphics[keepaspectratio=true,scale=0.16]{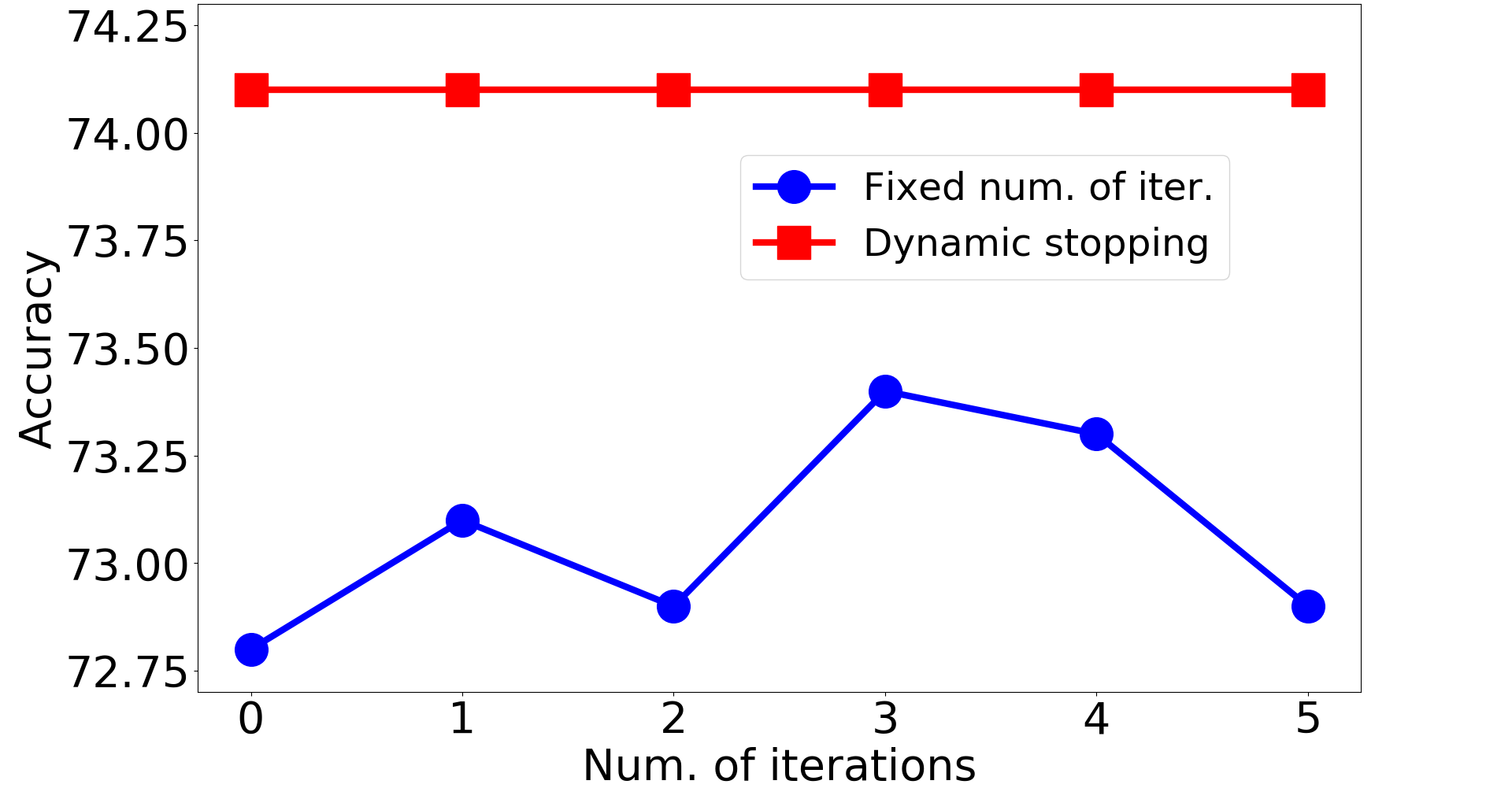}
    \caption{Performance comparison (i.e., test accuracy in \%) of two different stopping strategies on Citeseer.}
%   \label{fig:citeseer_stopping}
\end{figure}

% Graph visualization
\subsection{Graph Visualization}\label{sec:graph_viz}

Here, we visualize the graph structures (i.e., $\vec{A}^{(t)}$) learned by \alg. 
As we can see, compared to the initial graph structures, \alg mainly forms graph structures within the same class of nodes, which complement the initial graph structure.
This is as expected because $\vec{A}^{(t)}$ is computed based on the updated node embeddings that are supposed to capture certain node label information.

\begin{figure}[!tbh]
%   \vspace{-5mm}
  \centerline{
  \subfloat[t][Initial graph ($\vec{A}^{(0)}$)]{%
    \includegraphics[keepaspectratio=true,scale=0.16]{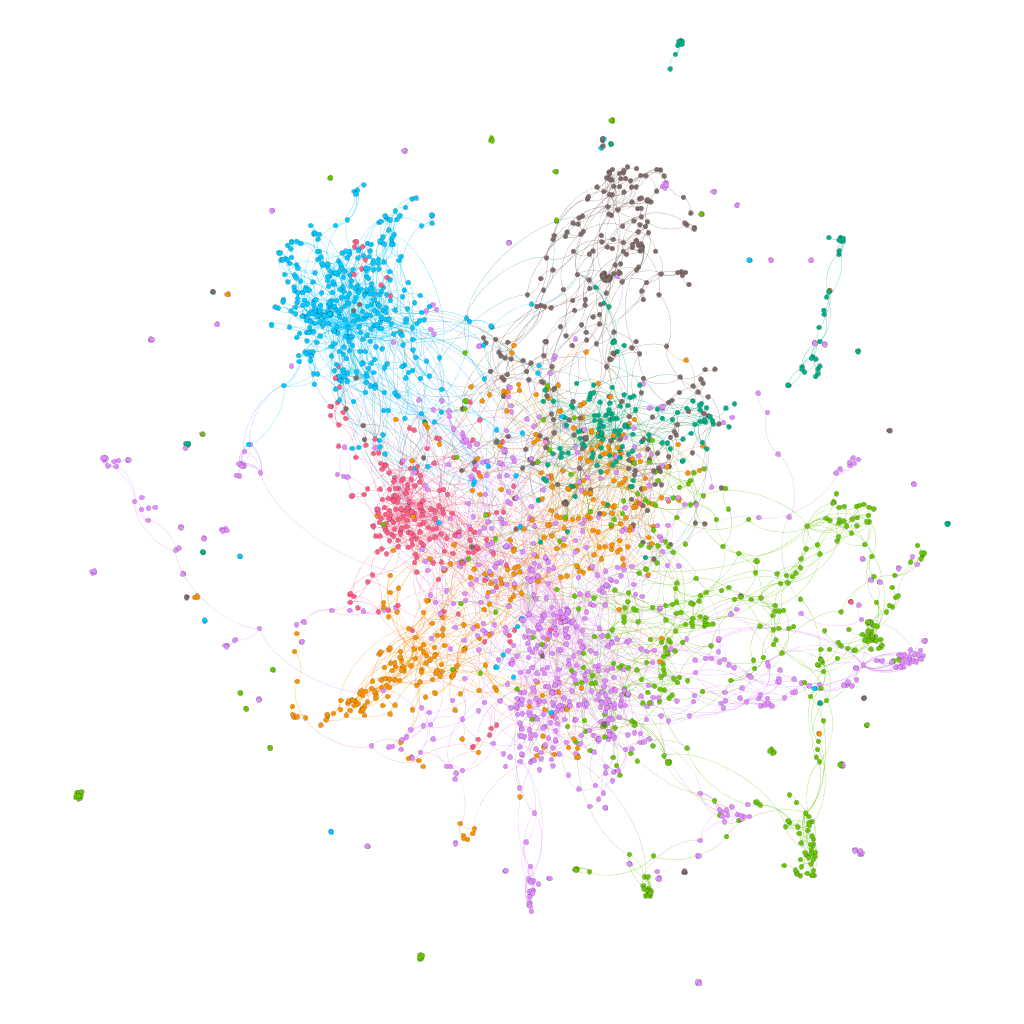}
	\label{fig:cora_init_graph}
	}
	  \hspace{0.5in}
  \subfloat[t][Learned graph ($\vec{A}^{(t)}$)]{%
    \includegraphics[keepaspectratio=true,scale=0.16]{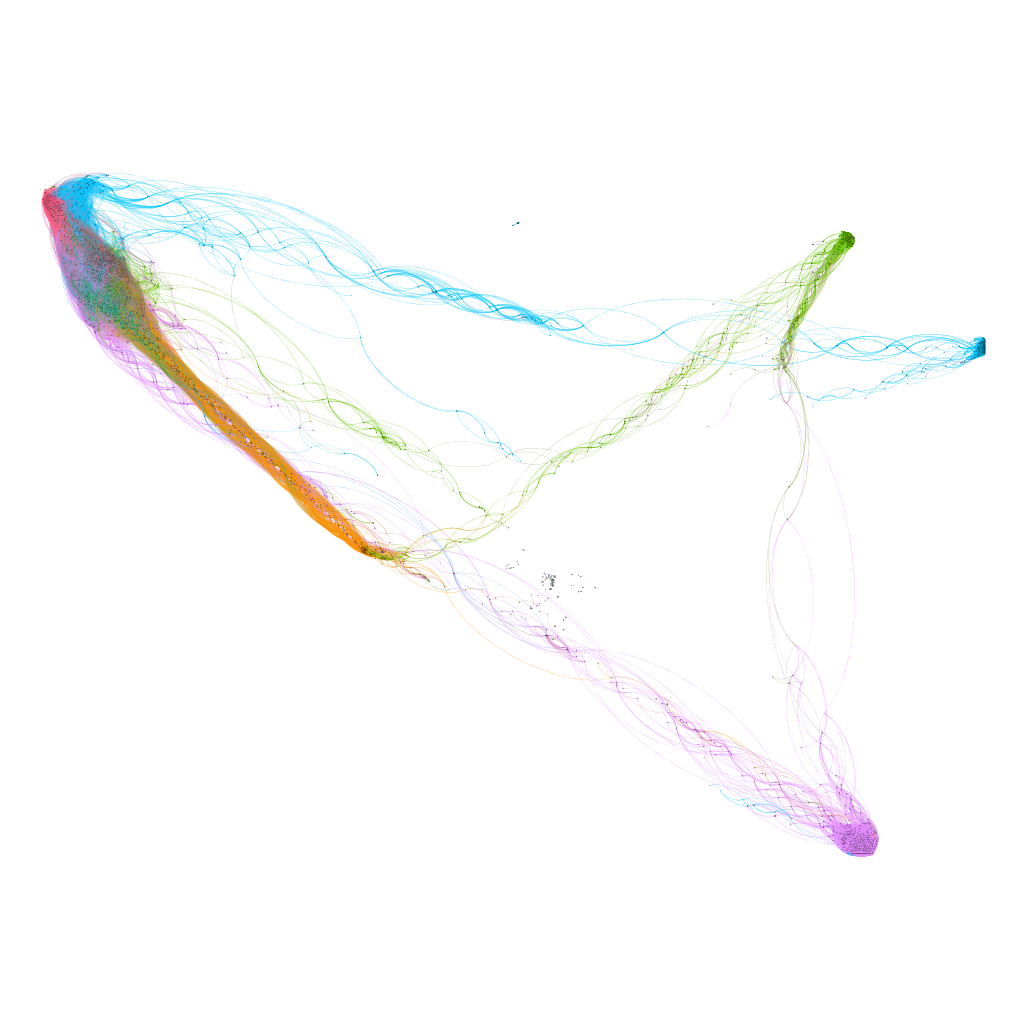}
	\label{fig:cora_idgl_graph}
	}
 }
%   \vspace{-0.1in}
    \caption{
    Visualization of the initial graph and the learned graph on Cora. Colors indicate different node labels.}
  \label{fig:cora_graph_viz}
%   \vspace{-5mm}
\end{figure}

\begin{figure}[!tbh]
  \centerline{
  \subfloat[t][kNN graph ($\vec{A}^{(0)}$)]{%
    \includegraphics[keepaspectratio=true,scale=0.16]{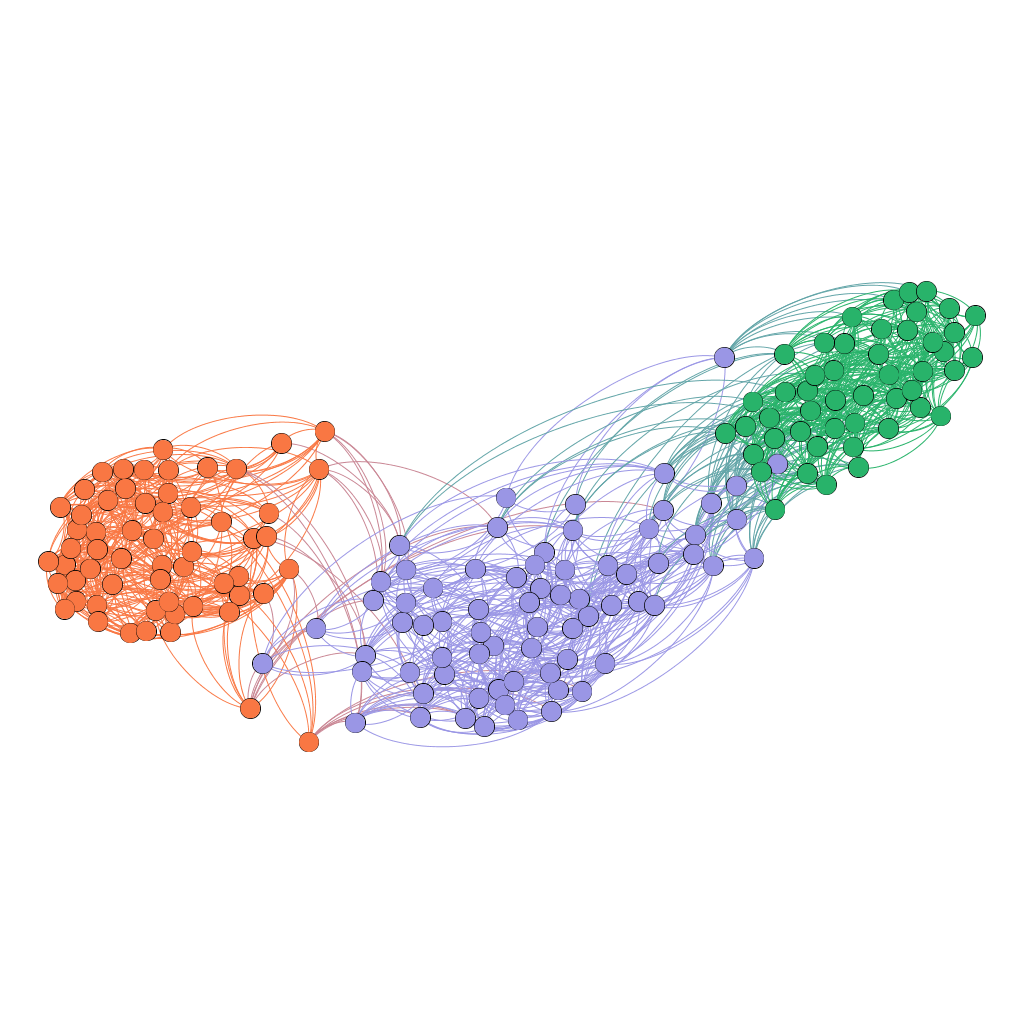}
	\label{fig:wine_knn_graph}
	}
\hspace{0.5in}
  \subfloat[t][Learned graph ($\vec{A}^{(t)}$)]{%
    \includegraphics[keepaspectratio=true,scale=0.16]{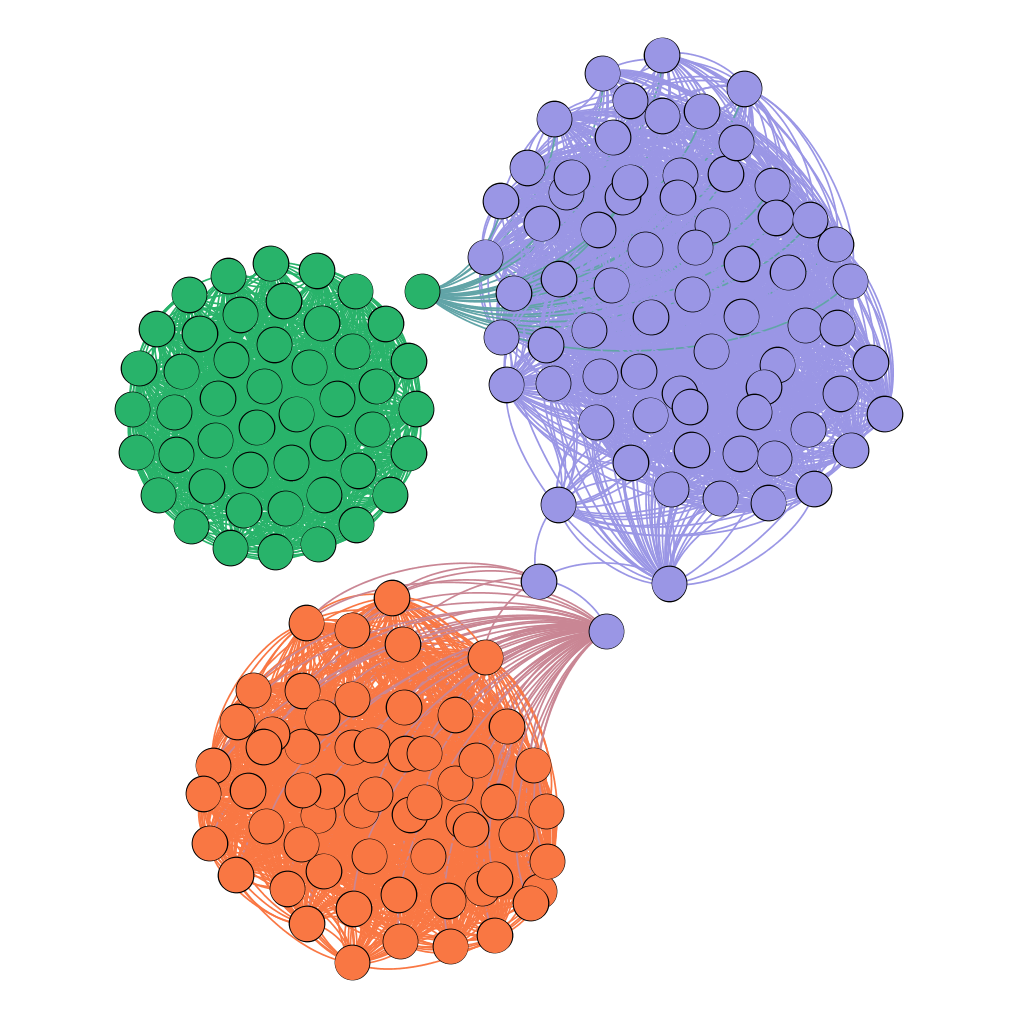}
	\label{fig:wine_idgl_graph}
	}
 }
    \caption{Visualization of the kNN graph and the learned graph on Wine. Colors indicate different node labels.}
  \label{fig:wine_graph_viz}
\end{figure}

\subsection{Hyperparameter Analysis}\label{sec:hyperparam_analysis}

A hyperparameter $\lambda$ is used to balance the trade-off between using the learned graph structure and the initial (or kNN) graph structure. 
In~\cref{table:effect_of_lambda}, we show the results of using different values of $\lambda$ on Cora.

We also study the effect of the hyperparameter $s$ (i.e., the number of anchors in \salg). As shown in~\cref{table:effect_of_s}, lower value of $s$ can degrade the performance of \salg whereas after certain optimal value, further increasing the number of anchors might not help the performance.

\begin{table*}[!htb]
% \vspace{-4mm}
\caption{Test scores ($\pm$ standard deviation) with different values of $\lambda$ on the Cora data.}
\label{table:effect_of_lambda}
\centering
\begin{tabular}{lllllll}
\hline
  Methods / $\lambda$ & \vline & 0.9 & 0.8 & 0.7 & 0.6 & 0.5 \\
    \hline
  \alg & \vline & 83.6 (0.4) & \textbf{84.5 (0.3)} & 83.9 (0.3) &  82.4 (0.1) & 80.9 (0.2) \\
  \salg & \vline &83.2 (0.4) & \textbf{84.4 (0.2)} &83.5 (0.6)  & 82.9 (0.4)  &54.6 (32.3)\\
 \hline
\end{tabular}
% \vspace{-4mm}
\end{table*}

\begin{table*}[!htb]
% \vspace{-4mm}
\addtolength{\tabcolsep}{-1pt}
\caption{Test scores ($\pm$ standard deviation) with different values of $s$ for \salg on the Cora and Pubmed data.}
\label{table:effect_of_s}
\centering
\begin{tabular}{llllllll}
\hline
  Methods / $s$ & \vline & 1,600 & 1,300 & 1,000 & 700 & 400 & 100 \\
    \hline
  Cora & \vline & 84.0 (0.4)  &84.1 (0.5) & \textbf{84.4 (0.2)} & 83.8 (0.2) & 58.7 (30.5) & 38.3 (25.9) \\
  Pubmed & \vline & 82.7 (0.2)  & \textbf{83.0 (0.4)} & 82.7 (0.4) & \textbf{83.0 (0.2)} &82.7 (0.3)  & 82.4 (0.5)\\
 \hline
\end{tabular}
% \vspace{-4mm}
\end{table*}

\section{Details on Experimental Setup}

\subsection{Data Statistics}\label{sec:data_statistics}

\cref{table:data_statistics} shows the data statistics of the nine benchmarks used in our experiments.

% \begin{table*}[!htb]
% % \vspace{-4mm}
% \caption{Data statistics.}
% \label{table:data_statistics}
% \centering
% \begin{tabular}{lllll}
% \hline
%   Benchmarks & \vline & Train/Dev/Test & Task & Setting\\
%   \hline
%   Cora & \vline &140/500/1,000
% &node classification & transductive\\ 
% Citeseer & \vline &120/500/1,000
% &node classification & transductive\\ 
% Pubmed & \vline & 60/500/1000
% &node classification & transductive\\ 
% Wine & \vline &10/20/158
% &node classification & transductive\\ 
% Cancer & \vline &10/20/539
% &node classification & transductive\\ 
% Digits & \vline &50/100/1,647
% &node classification & transductive\\ 
% 20News & \vline &7,919/3,395/7,532
% &graph classification & inductive\\ 
% MRD & \vline &3,003/1,001/1,002
% &graph regression & inductive\\ 

%  \hline
% \end{tabular}
% % \vspace{-4mm}
% \end{table*}

\begin{table}[!htb]
% \vspace{-4mm}
% \addtolength{\tabcolsep}{-0.8pt}
\caption{Data statistics. (clf. indicates classification and reg. indicates regression.)}
\label{table:data_statistics}
\centering
\scalebox{0.9}{
\begin{tabular}{lllllll}
\hline
  Benchmarks & \vline & \#Nodes & \#Edges & Train/Dev/Test & Task & Setting\\
  \hline
  Cora & \vline &2,708 (1 graph) &5,429&140/500/1,000
&node clf. & transductive\\ 
Citeseer & \vline &3,327 (1 graph)&4,732&120/500/1,000
&node clf. & transductive\\ 
Pubmed & \vline &19,717 (1 graph)&44,338& 60/500/1,000
&node clf. & transductive\\ 
ogbn-arxiv & \vline &169,343 (1 graph) &1,166,243& 90,941/29,799/48,603
&node clf. & transductive\\ 
Wine & \vline &178 (1 graph)&N/A&10/20/158
&node clf. & transductive\\ 
Cancer & \vline &569 (1 graph)&N/A&10/20/539
&node clf. & transductive\\ 
Digits & \vline &1,797 (1 graph)&N/A&50/100/1,647
&node clf. & transductive\\ 
20News & \vline &317 (18,846 graphs)&N/A&7,919/3,395/7,532
&graph clf. & inductive\\ 
MRD & \vline &389 (5,006 graphs)&N/A&3,003/1,001/1,002
&graph reg. & inductive\\ 

 \hline
\end{tabular}
}
% \vspace{-4mm}
\end{table}

\subsection{Model Settings}\label{sec:model_settings}

\begin{table*}[!htb]
% \vspace{-4mm}
\caption{Hyperparameter for \alg on all benchmarks.}
\label{table:hyperparam}
\centering
\begin{tabular}{lllllllllllll}
\hline
  Benchmarks & \vline & $\lambda$ & $\eta$ & $\alpha$& $\beta$ & $\gamma$ & $k$ & $\epsilon$ & $m$ & $\delta$ &  $T$\\
  \hline
  Cora & \vline &0.8& 0.1& 0.2 &0.0&0.0 &\textrm{--}&0.0&4&4.0e-5 & 10\\ 
  Citeseer & \vline &0.6& 0.5& 0.4 &0.0&0.2 &\textrm{--}&0.3&1.0&1.0e-3 & 10\\
 Wine & \vline &0.8& 0.7& 0.1 &0.1&0.3 &20&0.75&1&1.0e-3 & 10\\
 Cancer & \vline &0.25& 0.1& 0.4 &0.2&0.1 &40&0.9&1&1.0e-3 & 10\\
 Digits & \vline &0.4& 0.1& 0.4 &0.1&0.0 &24&0.65&8&1.0e-4 & 10\\
%  20News10 & \vline &0.6& 0.2& 0.2 &0.1&0.1 &10&0.2&8&1e-3 & 10\\
 20News & \vline &0.1& 0.4& 0.5 &0.01&0.3 &950&0.3&12&8.0e-3 & 10\\
 MRD & \vline &0.5& 0.9& 0.2 &0.0&0.1 &350&0.4&5&4.0e-2 & 10\\
 \hline
\end{tabular}
% \vspace{-4mm}
\end{table*}

\begin{table*}[!htb]
% \vspace{-4mm}
\caption{Hyperparameter for \salg on all benchmarks.}
\label{table:anchor_hyperparam}
\addtolength{\tabcolsep}{-2pt}
\centering
\begin{tabular}{llllllllllllll}
\hline
  Benchmarks & \vline & $\lambda$ & $\eta$ & $\alpha$& $\beta$ & $\gamma$ & $k$ & $\epsilon$ & $m$ & $\delta$ &  $T$ & num./ratio of anchors\\
  \hline
  Cora & \vline & 0.8 &0.1 & 0.2 &0.0&0.1 &\textrm{--}&0.0&4& 8.5e-5&10 &1,000\\ 
  Citeseer & \vline &0.6&0.5 &0.5 &0.1&0.2 &\textrm{--}&0.2&4&2.0e-3& 10&1,400\\
  Pubmed & \vline &0.7&0.3 &0.0 &0.03&0.0 &\textrm{--}&0.1&6&8.0e-5 &10&700\\
  ogbn-arxiv & \vline &0.8&0.1 & 0.2&0.0&0.0 &\textrm{--}&0.9&1&1.0e-1 &2&300\\
 Wine & \vline &0.7 &0.7  &0.1  & 0.1& 0.3 &  20&0.75& 1 & 1.0e-3 & 10 &200\\
 Cancer & \vline & 0.25& 0.1& 0.0 & 0.0& 0.0 & 40 &0.9 & 4& 8.0e-4 & 10 & 100\\
 Digits & \vline &0.3 &0.3  & 0.4  & 0.1& 0.0& 24& 0.65& 8& 1.0e-4 & 10 &1,500\\
 20News & \vline & 0.1& 0.3 & 0.4  & 0.0& 0.3 &950 & 0.4& 12 & 1.0e-2 & 10 & 0.4\\
 MRD & \vline & 0.5 & 0.75 & 0.2  & 0.0& 0.0 &400 & 0.7&4 & 3.0e-2 & 10 &0.4\\
 \hline
\end{tabular}
% \vspace{-4mm}
\end{table*}

In all our experiments, we apply a dropout ratio of 0.5 after GCN layers except for the output GCN layer.
During the iterative learning procedure, we also apply a dropout ratio of 0.5 after the intermediate GCN layer, except for Citeseer (no dropout) and Digits (0.3 dropout).
For experiments on text benchmarks,
we keep and fix the 300-dim GloVe vectors for words that appear more than 10 times in the dataset.
For long documents, for the sake of efficiency, we cut the text length to maximum 1,000 words.
We apply a dropout ratio of 0.5 after word embedding layers and BiLSTM layers.
The batch size is set to 16.
And the hidden size is set to 128 and 64 for 20News and MRD, respectively.
For all other benchmarks, the hidden size is set to 16 to follow the original GCN paper.
For the text benchmarks, we apply a BiLSTM to a sequence of word embeddings.
The concatenation of the last forward and backward hidden states of the BiLSTM is used as the initial node features.
We use Adam \citep{kingma2014adam} as the optimizer.
For the text benchmarks, we set the learning rate to 1e-3. 
For all other benchmarks, we set the learning rate to 0.01 and apply L2 norm regularization with weight decay set to 5e-4.
As for \salg, we set the number of anchors as a hyperparameter in transductive experiments, while in inductive experiments, we set the ratio of anchors (proportional to the graph size) as a hyperparameter.
In ~\cref{table:hyperparam} and~\cref{table:anchor_hyperparam}, we show the hyperparameters for \alg and \salg on all benchmarks, respectively.
All hyperparameters are tuned on the development set.

\end{document}